\documentclass[letterpaper, 10 pt, conference]{ieeeconf}
\IEEEoverridecommandlockouts
\usepackage[utf8]{inputenc}
\usepackage{amsmath}
\usepackage{amssymb}
\usepackage{graphicx}
\usepackage{subcaption}
\usepackage{color}
\let\labelindent\relax
\usepackage[shortlabels]{enumitem}
\usepackage{hyperref}
\usepackage{bbm}
\usepackage{tabu}
\usepackage{booktabs}
\usepackage{multirow}
\usepackage{float}
\usepackage{svg}
\usepackage
[backend=bibtex,
bibstyle=ieee,
citestyle=numeric,
sortcites,
natbib=true,
doi=false,
isbn=false,
url=true,
hyperref=true,
sorting=nyt,
eprint=false,
maxbibnames=99]{biblatex}

\newcommand{\norm}[2][]{\left\lVert#2\right\rVert_{#1}}

\newcommand{\argmin}{\mathop{\mathrm{arg\,min}}}

\usepackage{mathtools}

\DeclarePairedDelimiter{\abs}{|}{|}
\DeclarePairedDelimiter{\set}{\{}{\}}
\DeclarePairedDelimiter{\paren}{(}{)}
\newcommand{\mmatrix}[1]{\begin{bmatrix}#1\end{bmatrix}}

\renewcommand{\vec}[1]{\mathbf{#1}}
\newcommand{\p}[1]{\paren*{#1}}

\newcommand{\suchthat}{\hspace{0.2em}|\hspace{0.2em}}

\usepackage{algorithm,algorithmicx}
\usepackage[noend]{algpseudocode}

\definecolor{Blue}{rgb}{0,0,1}
\definecolor{Orange}{rgb}{1,0.65,0}

\title{3D Force and Contact Estimation for a Soft-Bubble Visuotactile Sensor Using FEM}
\author{Jing-Chen Peng$^{1}$, Shaoxiong Yao$^{1}$, and Kris Hauser$^{1}$, Senior Member, IEEE
\thanks{$^1$: Department of Computer Science at the University of Illinois at Urbana-Champaign, Urbana, Illinois, USA.      
    {\tt\small(jcpeng2,syao16,kkhauser)\@illinois.edu} This work is partially supported by NIFA/USDA Award \# 2020-67021-32799.}%
}
\begin{document}

\maketitle

\begin{abstract}

Soft-bubble tactile sensors have the potential to capture dense contact and force information across a large contact surface.
However, it is difficult to extract contact forces directly from observing the bubble surface because local contacts change the global surface shape significantly due to membrane mechanics and air pressure.  
This paper presents a model-based method of reconstructing dense contact forces from the bubble sensor's internal RGBD camera and air pressure sensor.
We present a finite element model of the force response of the bubble sensor that uses a linear plane stress approximation that only requires calibrating 3 variables. Our method is shown to reconstruct normal and shear forces significantly more accurately than the state-of-the-art, with comparable accuracy for detecting the contact patch, and with very little calibration data. 

\end{abstract}

\begin{figure}[tbp]
    \centering
    \includegraphics[width=\linewidth]{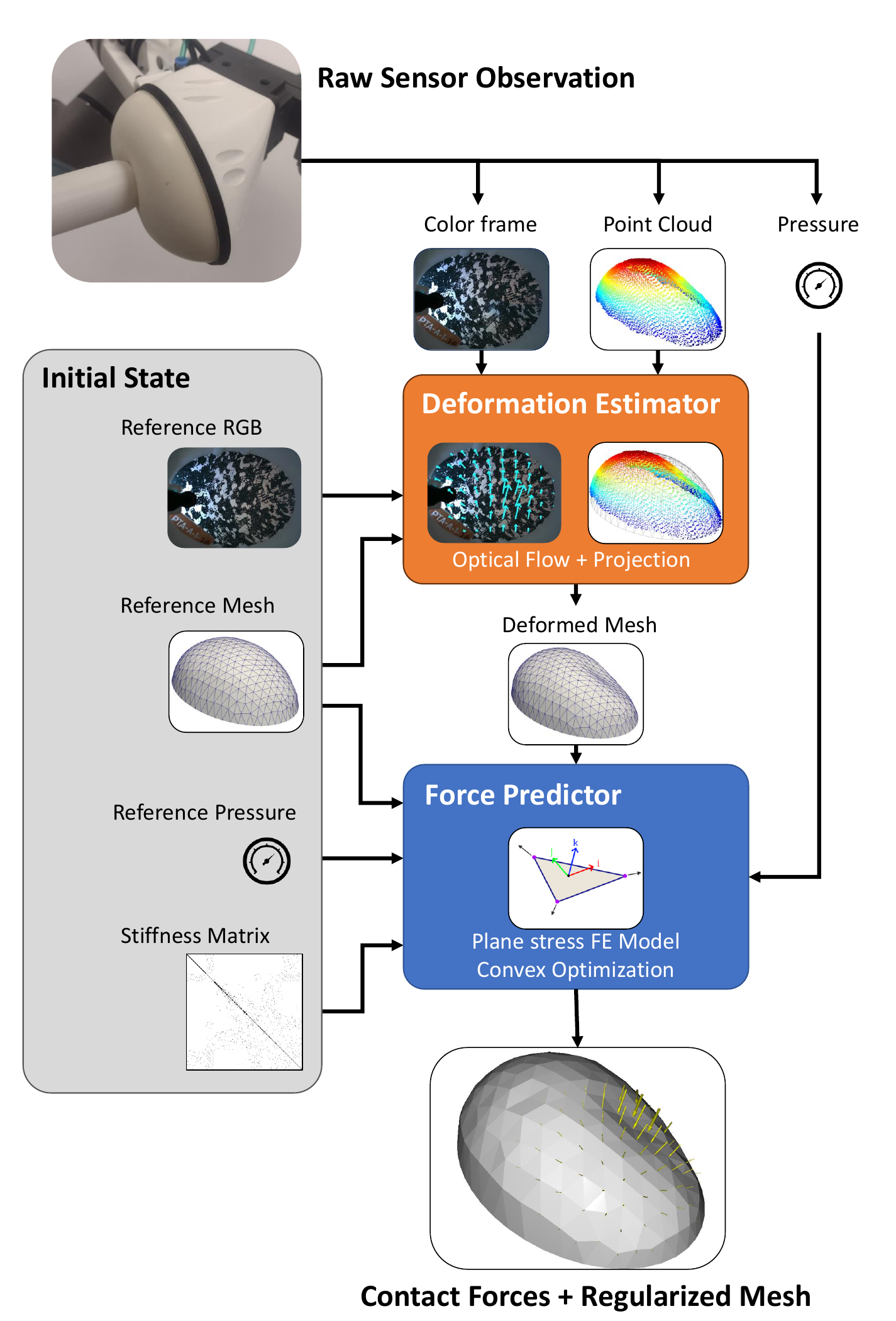}
    \caption{
    A summary of the force estimation pipeline.
    An RGB image and point cloud from the bubble sensor are used to estimate the 3D mesh deformation relative to a reference configuration.
    We construct a finite element model of the bubble to predict forces from observed deformations and pressure changes.
    A convex optimization problem combines the physics model and camera observations, giving contact force estimates and regularizing the raw observations.
    }
    \vspace{-0.1in}
    \label{fig:intro}
    \vspace{-0.2in}
\end{figure}

\section{Introduction}

Tactile information is needed for robots to perform safe and precise manipulation in emerging domains like healthcare~\cite{erichson2018icra} and industrial assembly~\cite{tang2023rss}.  Recent years have seen a rapid growth in interest in vision-based tactile sensors like Gelsight~\cite{gelsight} and Punyo~\cite{alspach2019softbubble}, which can provide high-resolution contact region observations by sensing the deformation internally via embedded cameras.  Vision-based tactile sensors can be used to perceive objects' material properties~\cite{yuan2017icra} and perform efficient manipulation~\cite{she2021ijrr}.  
However, inferring accurate contact information remains a significant challenge.  Model-based methods have been applied to contact force estimation for elastomer sensors~\cite{gelsight} and normal force estimation for soft-bubble sensors~\cite{punyo_force}.  Learning-based methods have also been used to address estimation problems for such sensors ~\cite{yuan2017icra,funk2023highresolution,lambeta2021pytouch}, but typically require a large number of observations and fail to generalize out of distribution.

\begin{figure}[tbp]
    \centering
    \includegraphics[width=\linewidth]{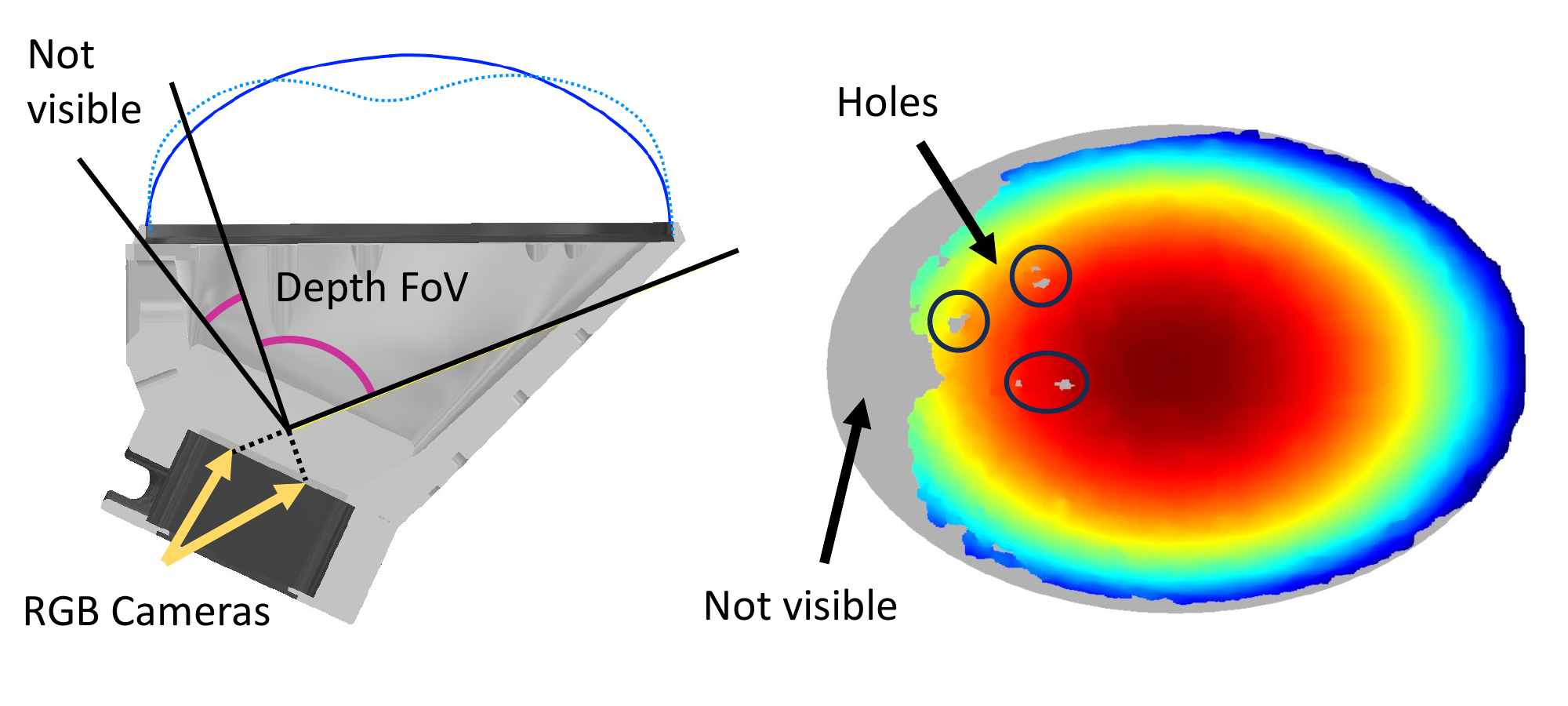}
    \caption{
    Left: A schematic of the Punyo soft-bubble sensor used for our experiments, showing the camera layout and visible regions.
    Right: A typical point cloud reading, colored by Z coordinate. [Best viewed in color]
    }
    \label{fig:punyo_fig}
    \vspace{-0.2in}
\end{figure}

In this work, we present a finite-element (FE) based model of a \textit{soft-bubble} vision-based tactile sensor~\cite{alspach2019softbubble} to produce dense contact force and patch estimation using the sensor's internal camera and pressure sensor.
A schematic of one such sensor is given in \autoref{fig:punyo_fig}:
A depth camera provides 3D deformation information about an elastic membrane, pressurized with air.
Simple image differences are quite poor for computing the contact patch~\cite{lambeta2021pytouch} because the bubble's deformation is not restricted to the contact area.
Our work is an evolution of Kuppuswamy et al.~\cite{punyo_force}, who proposed an FE model based on the curvature of the bubble. However, they assume a frictionless bubble and their model can only estimate normal contact forces. In contrast, our method can estimate general 3D contact forces (i.e., shear). We use depth and optical flow to approximate the deformed mesh shape and a full 3D FE model of the pressure distribution and membrane stresses.  We formulate contact force estimation as a convex optimization problem maximizing the mesh deformation likelihood with
$L_1$ regularization on contact forces.
The force estimator has 3 parameters and can be calibrated well using as few as 6 touch trajectories. Experiments show that our method leads to $36\%$ reduced force estimate error compared to the prior state-of-the-art~\cite{punyo_force} and comparable accuracy in estimating the contact patch.
Code will be made available upon paper acceptance.

\section{Related Work}
Recently, there has been an increasing interest in designing novel tactile sensors~\cite{sensors2018}.  Vision-based tactile sensors such as the Gelsight~\cite{gelsight}, TacTip~\cite{tactip} and DIGIT~\cite{lambeta2020digit} can provide high-resolution contact observations and have been proven useful in manipulation tasks~\cite{Shah2021}.
The Punyo \textit{soft-bubble} sensor~\cite{alspach2019softbubble} is an air-inflated bubble with an internally mounted RGBD camera, which has a large sensing area and is resilient to large contact forces.
A schematic of the sensor is shown in \autoref{fig:punyo_fig}.
Due to the richness of sensor data coming from internal cameras, these devices have been used for identifying objects through touch~\cite{yuan2017icra} and localization of grasped objects~\cite{she2021ijrr,lambeta2020digit}.  
A more fundamental task is to estimate external contact force and locations across the sensor surface from the observed deformations.  By bridging raw observations to physical quantities, such methods enable physics-based reasoning that can be employed in models for higher level tasks like object pose detection and recognition. Moreover, it is likely that force estimators could be calibrated for a particular sensor with less data than learning estimators for each desired higher-order task.

Existing methods for force estimation can be divided into model-based and data-driven methods.  Early work by Ito et al.~\cite{sensors2014} used the kinematics of dots on the sensor surface to categorize dots as slipping, sticking, or in free space.  The initial version of the GelSight~\cite{gelsight} has a similar dot pattern that provides high-resolution deformation tracking.  To estimate force information, Ma et al. apply inverse FEM to estimate contact forces on the GelSlim 2.0 sensor~\cite{gelsight_FEM}.  They demonstrated that linear FEM can sufficiently capture the relation between deformation and contact forces in a gel-type visuotactile sensor.
However, their method models the deforming material using solid elements, which are not suited for modeling a membrane-like sensor architecture.
More recently, data-driven methods have been developed to learn a mapping from visual to contact information.
Wang et al.~\cite{Wang2021GelSightWM} use a small neural network to learn a mapping from RGB pixels to surface normals and reconstruct the depth map by solving Poisson's equation.
DenseTact 2.0~\cite{do2023densetact} uses an encoder-decoder network to directly map visual observation to contact deformation and forces.
Oller et al. used a PointNet style approach to estimate deformation and external force~\cite{oller2023corl}.
Funk et al.~\cite{funk2023highresolution} used a U-Net structure to directly estimate contact forces from a raw tactile image.  
While learning-based approaches have seen success, they are very data hungry and time-consuming: For example, \cite{funk2023highresolution} trains on 277325 samples across 12 different indenting shapes, \cite{do2023densetact} took more than 10 GPU hours in training.
In contrast, we are interested in developing data-efficient methods for force estimation.

For the soft-bubble sensor we studied in this paper, estimating contact information from vision is challenging because contact induces deformation across the entire membrane.  To address this challenge, Kuppuswamy et al. use a FE membrane model to simulate the behavior of a soft-bubble sensor and solve the inverse problem to compute contact patches and contact forces~\cite{punyo_force}. However, their model assumes a frictionless bubble surface, ignoring shear forces.  Our model extends their method to 3D mesh deformations and is able to capture general 3D contact forces, improving the contact force estimation by $36\%$ on average while retaining comparable contact patch estimation.

\section{Model Based Force Estimation}
Our model is based on a triangular mesh discretization of the bubble's surface and estimates contact forces at each node on the mesh.
We will refer to the entire mesh as $M$, an individual triangle as $\Delta$, and the total number of nodes as $\abs{M}$.
$\partial M$ will refer to the set of nodes that are on the boundary of $M$.
In order to estimate contact forces, our model performs three steps:
First, the 3D deformation of the bubble's surface is estimated from RGBD camera readings and interpolated to the mesh.
Second, a stiffness model is constructed, relating the mesh points' displacements to forces at each mesh node.
Finally, a convex optimization problem is solved to estimate contact forces while regularizing for noise introduced from the camera observation and interpolation process.
An overview of the entire process is given in \autoref{fig:intro}.

\subsection{3D Deformation estimation from RGBD images}
\label{sec:deform_est}
To estimate the 3D motion of points on the bubble surface, we combine the depth camera's point cloud with optical flow from the RGB image.
We calibrate the camera and indicate 3D to image space projection as $\mathcal{P}(\cdot): \mathbb{R}^3 \to \mathbb{R}^2$.  Given a depth image $D$, we can also project each pixel to 3D using the interpolated depth map $\mathcal{S}(\cdot): \mathbb{R}^2 \to \mathbb{R}^3$.

Our model computes deformations relative to a reference configuration, taken when the sensor is inflated and has no external contacts.
This consists of a reference mesh $M_{ref}$ with nodes $\vec{x}_{ref}$ representing the bubble's undeformed pose, a reference RGB image for computing optical flow, and a reference pressure for computing pressure differences.

To compute the current position $\vec{x}_{cur} \in \mathbb{R}^3$ of a node in the reference mesh $\vec{x}_{ref} \in \mathbb{R}^3$, 
we first project $\vec{x}_{ref}$ into image space, computing $\vec{r} = P(\vec{x}_{ref})$.
Then, we use an optical flow map $\mathcal{F}(\cdot): \mathbb{R}^2 \to \mathbb{R}^2$ to displace $\vec{r}$: $\vec{r}' = \vec{r} + \mathcal{F}(\vec{r})$.
Finally, we project the displaced pixel into 3D space using $\mathcal{S}$.
In summary, the complete mapping from an undeformed node to its deformed position is
\begin{equation}
    \begin{split}
        \vec{x}_{cur} = \mathcal{S}(\mathcal{P}(\vec{x}_{ref}) + \mathcal{F}(\mathcal{P}(\vec{x}_{ref})).
    \end{split}
\end{equation}

$\mathcal{F}$ is computed by interpolating outputs of Farneback's dense flow algorithm~\cite{farneback_flow}~\cite{opencv_library}.
We use optical flow to find pixel correspondences because the \textit{soft-bubble} has a textured pattern on its internal surface~\cite{alspach2019softbubble}.
The interpolation of the depth map is performed using SciPy's barycentric interpolation~\cite{2020SciPy-NMeth}, which also helps remove holes in the depth map and fill in occluded regions (See \autoref{fig:punyo_fig}).

\subsection{Membrane Model}
The goal of the membrane model component is to establish a relationship between deformation of the bubble and their resulting forces.
We model the bubble sensor as a homogeneous thin membrane, similar to \cite{punyo_force}.
Our model considers three types of forces acting on each element of the bubble:
Tension forces from neighboring elements,
external forces from contact with the environment,
and pressure force from the air inside.
We consider the bubble deformation to be quasi-static, and solve for static equilibrium:
\begin{equation}
    \begin{split}
        F_{tension} + F_{pressure} + F_{external} = 0.
    \end{split}
    \label{eq:equilibrium}
\end{equation}
We further linearize \autoref{eq:equilibrium} about a fixed reference configuration of the bubble, defined in \autoref{sec:deform_est}:
\begin{equation}
    \begin{split}
        \delta F_{tension} + \delta F_{pressure} + F_{external} = 0,
    \end{split}
    \label{eq:equilib_2}
\end{equation}
where $\delta F_{external} = F_{external}$ since the reference configuration has no external forces acting on it.

We impose these equilibrium conditions at every mesh vertex and solve for values of these quantities at those vertices.
For this discussion, $\delta F_i \in \mathbb{R}^{\abs{M} \times 3}$ will denote a matrix of stacked node forces, for each of the three force quantities.

We first lump the change in pressure force to each vertex to compute $\delta F_{pressure}$.  
We lump the area to each vertex as $\vec{a} \in \mathbb{R}^{\abs{M}} \suchthat a_i = \frac{1}{3} \sum \set{\text{area}(\Delta) \hspace{0.2em} \forall \hspace{0.2em} \Delta \in M \suchthat i \in \Delta}$.
Similarly, we compute outward area-weighted vertex normals $\vec{n}_i \in \mathbb{R}^{3}$ for each vertex $i$.
$\delta F_{pressure}$ can be computed as:
\begin{equation}
    \delta F_{pressure} = \delta p \mmatrix{
        {a_{1}}\vec{n}_{1}^T \\
        {a_{2}}\vec{n}_{2}^T \\
        \vdots \\
        {a_{\abs{M}}}\vec{n}_{\abs{M}}^T \\
    }
    \label{eq:pressure}
\end{equation}
where $\delta p$ is the difference between the observed pressure reading and the reference pressure reading.

$\delta F_{tension}$ is computed by assuming linear elasticity.
We characterize the material by its Young's modulus $E$ and Poisson ratio $\nu$.
Following Reissner-Minlin plate theory, we assume that the membrane thickness does not change during deformation and that the out of plane stress is zero (\cite{fem_book}, Section 5.2).
Additionally, we assume the bubble's bending stiffness is zero because the membrane is very thin (0.65mm) compared to its radius of curvature.

\begin{figure}
    \centering
    \includegraphics[width=0.8\linewidth]{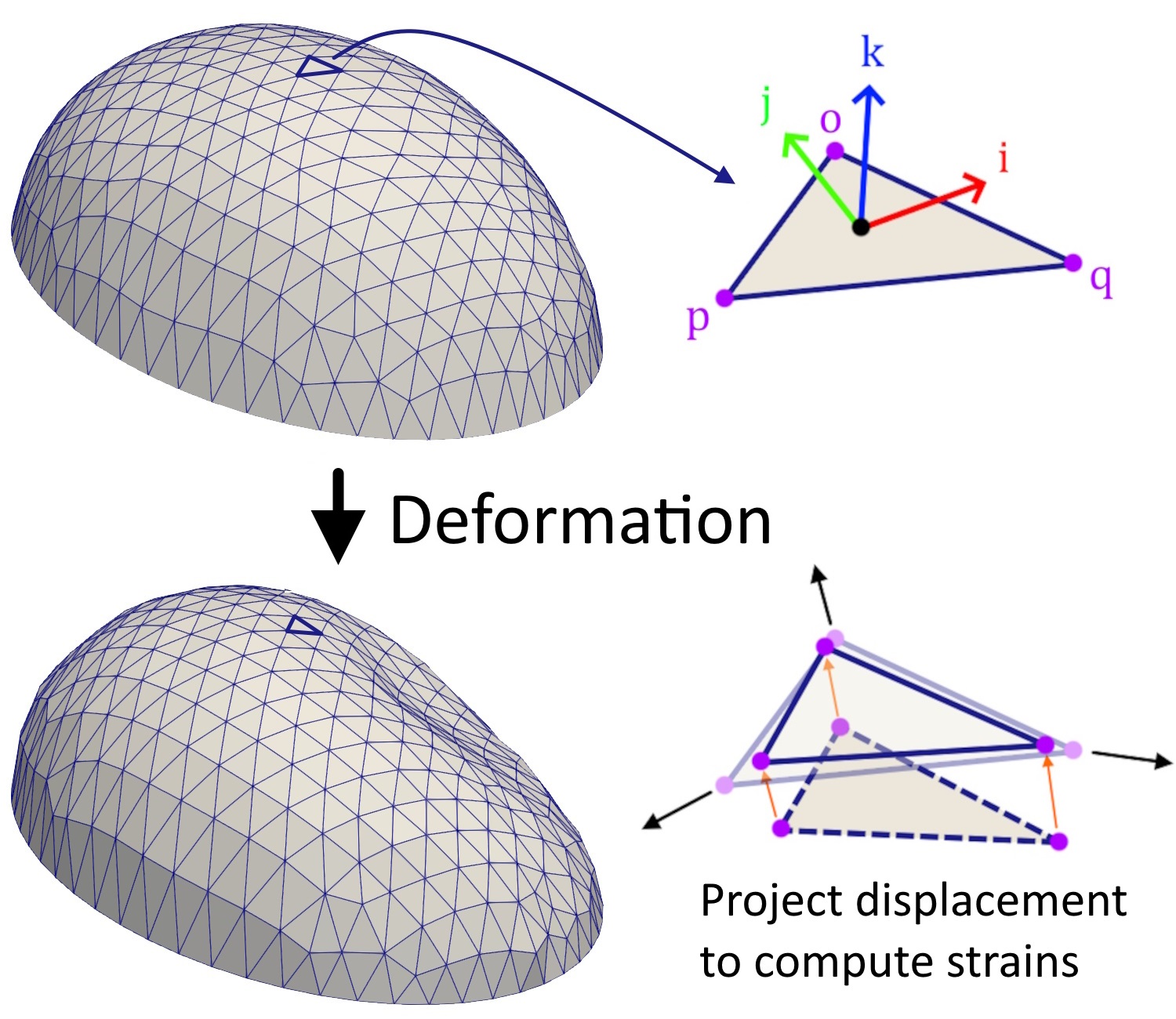}
    \caption{An illustration of the FEM setup.
    A local coordinate frame is computed for each triangle, and displacements are projected back to the local $ij$ plane to compute strains. [Best viewed in color]}
    \label{fig:triangle_project}
    \vspace{-0.2in}
\end{figure}

For a point on the mesh surface, we define the local coordinate system $ijk$ such that $i$ and $j$ are tangent to the surface and $k$ is perpendicular to the surface.
(Refer to \autoref{fig:triangle_project}.)
Let $\sigma_{ab}$ refer to element $(a,b)$ of the stress tensor $\sigma$.
$\varepsilon_{xx}$ refers to the normal strain in direction $x$, and $\gamma_{xy}$ refers to the shear strain in plane $xy$.
From the frame of reference of such a point, the linear elasticity equations reduce to:
\begin{equation}
    \begin{split}
        \varepsilon_{ii} = \frac{\sigma_{ii}}{E} - &\nu \frac{\sigma_j}{E}, \quad\quad
        \varepsilon_{jj}  = \frac{\sigma_{jj}}{E} - \nu \frac{\sigma_i}{E}, \\
        \gamma_{ij} & = \frac{2(1+\nu)}{E}\sigma_{ij}, \\
    \end{split}
\end{equation}
where the assumptions of no thickness change and no bending stiffness give additional constraints
\begin{equation}
    \varepsilon_{kk} = \sigma_{kk} = \sigma_{jk} = \sigma_{ik} = 0. \\
\end{equation}
These equations are applied on a triangular mesh, treating each triangle individually as a flat surface, following the procedure in \cite{fem_book} (Section 6.2.13).

As part of this simplification, when considering displacements of points in 3D, we project them onto the triangle plane for the purposes of calculating strain within each individual triangle.
Let $B$ be the partial derivative matrix for a linear triangle element in 2D, such that
\begin{equation}
    \mmatrix{\varepsilon_{ii} \\ \varepsilon_{jj} \\ \gamma_{ij}} = B \mmatrix{\vec{u}_{o,2D}^T & \vec{u}_{p,2D}^T & \vec{u}_{q,2D}^T}^T,
\end{equation}
where $\vec{u}_{v, 2D}$ refers to the 2D displacement of vertex $v$.
For our simplification, we set
\begin{equation}
    \vec{u}_{v, 2D} = \text{project}(\vec{u}_v) = \mmatrix{\hat{i} & \hat{j}}^T \vec{u}_v,
\end{equation}
where $\vec{u}_v$ refers to the observed 3D displacement of vertex $v$:
\begin{equation}
    \vec{u}_v = \vec{x}_{v, cur} - \vec{x}_{v, ref}
\end{equation}
Refer to \autoref{fig:triangle_project}.
The 2D computed node forces are translated back to 3D in an analogous fashion:
\begin{equation}
    \vec{f}_{v} = \mmatrix{\hat{i} & \hat{j}} \vec{f}_{v, 2D}.
\end{equation}

Following the standard FEM assembly procedure, we get a linear map $K$ between node displacements and node tension force changes:
\begin{equation}
    \delta F_{tension} = K\p{\mmatrix{\vec{u}_1^T & \vec{u}_2^T & \dots & \vec{u}_{|M|}^T}} = K(\vec{U})
    \label{eq:stiffness}
\end{equation}
where $\vec{U} \in \mathbb{R}^{3\abs{M}}$ denotes a vector of stacked node displacements.

Substituting \autoref{eq:pressure} and \autoref{eq:stiffness}, in \autoref{eq:equilib_2}, we obtain an expression for the unknown contact force $F_{external}$ as a function of the observed displacements $\vec{U}$ and pressure change $\delta p$:
\begin{equation}
    F_{external}(\vec{U}, \delta p) = -\p{K(\vec{U}) + \delta p \mmatrix{
        {a_{1}}\vec{n}_{1}^T \\
        {a_{2}}\vec{n}_{2}^T \\
        \vdots \\
        {a_{\abs{M}}}\vec{n}_{\abs{M}}^T \\
    }}
    \label{eq:f_ext}
\end{equation}

To compute contact pressures $P \in \mathbb{R}^{\abs{M} \times 3}$ at each node, we divide the computed node force by the area at each vertex:
\begin{equation}
    P_{contact} = diag(\vec{a})^{-1}
    F_{external}.
\end{equation}
We additionally define the continuous pressure distribution $\mathcal{P}_{contact}$ by barycentric interpolation of $P_{contact}$ within each triangular face.

The total contact force $\vec{f}_{net}$ is computed by summing the node forces across the mesh surface, excluding the boundary:
\begin{equation}
    \vec{f}_{net} = \sum_{v \in M \setminus \partial M} \p{F_{external}}_{v}
    \label{eq:force}
\end{equation}

\subsection{Regularization}
Although \autoref{eq:f_ext} allows direct computation of contact forces from the observed node displacements in theory, in practice noise from the camera observation and the non-physical nature of the interpolation scheme in \autoref{sec:deform_est} produce strong artifacts in the force prediction.
We introduce a term  $\delta \vec{U} \in \mathbb{R}^{\abs{M} \times 3}$ to correct for noise in the raw observations, and formulate a group lasso optimization problem:
\begin{equation}
    \begin{split}
        \argmin_\vec{\delta U} \quad &  \norm[2,1; W_f]{F_{external}(\vec{U} + \delta \vec{U})} + \norm[2; W_u]{\delta \vec{U}}^2 \\
        \text{s.t.} \quad & \vec{u}_i = 0 \text{ if vertex } i \in \partial M\\
    \end{split}
\end{equation}
where $\norm[W]{\cdot}$ refers to the $W$-weighted norm of a vector $\vec{x} \in \mathbb{R}^n$ or matrix $A \in \mathbb{R}^{r \times c}$:
\begin{equation}
    \begin{split}
        \norm[2,1; W]{A} & = \norm[1]{W\mmatrix{s_1 \\ s_2 \\ \vdots \\ s_r}}; \quad s_k = \sqrt{\sum_{j=1}^c A_{jk}^2} \\
        \norm[2, W]{\vec{x}}^2 & = \vec{x}^\top W\vec{x}.
    \end{split}
\end{equation}
Note that $F_{external}$ is a linear function of $\delta \vec{U}$, so the only variables in this optimization are the node displacements away from the observed mesh nodes.
Group lasso is chosen to encourage sparsity in the contact region detection, but without creating artifacts where forces will tend to concentrate in the coordinate directions.
We formulate and solve this optimization using CVXPY~\cite{diamond2016cvxpy}.

\subsection{Contact patch estimation}

Following \cite{punyo_force}, we use a fraction of the average contact pressure to estimate contact patches.
Define the inward normal contact pressure vector $\vec{p} \in \mathbb{R}^{\abs{M}}$ as
\begin{equation}
    p_i = -\p{\vec{n}_i}^T \cdot \p{P_{contact}}_{i}
\end{equation}
Vertex $v$ on the deformed bubble is considered in contact if
\begin{equation}
    p_v > \text{thresh}(\vec{p}) =\max\p{k_{const}, k_{linear}\frac{\vec{a}^T\vec{p}}{\norm[1]{\vec{a}}}},
\end{equation}
with $k_{const} = 2000 \text{ Pa}$ and $k_{linear} = 2$, where $\vec{a}^T\vec{p} / \norm[1]{\vec{a}}$ is the average contact pressure across the entire bubble.
$k_{const}$ is tuned to compensate for noise that is present in the model output due to random fluctuations from the sensor readings.

\begin{figure}[htpb]
    \centering
    \includegraphics[width=0.8\linewidth]{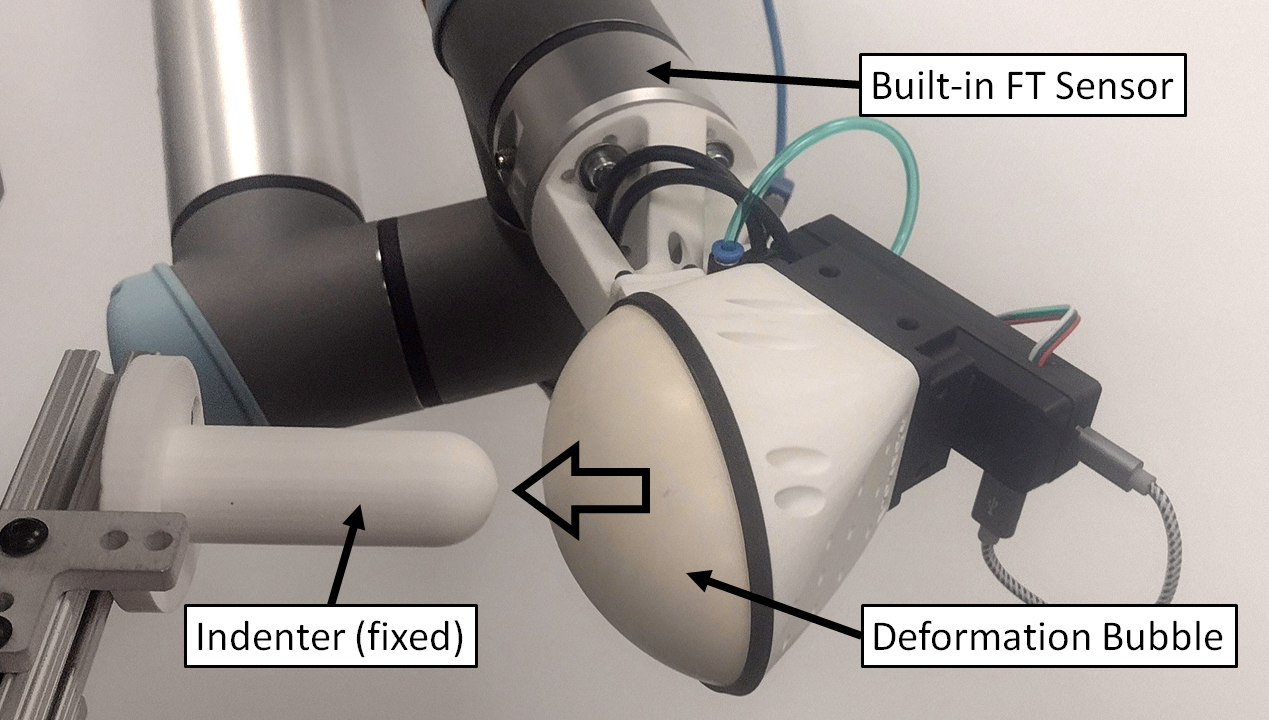}\\ 
    \includegraphics[width=0.8\linewidth]{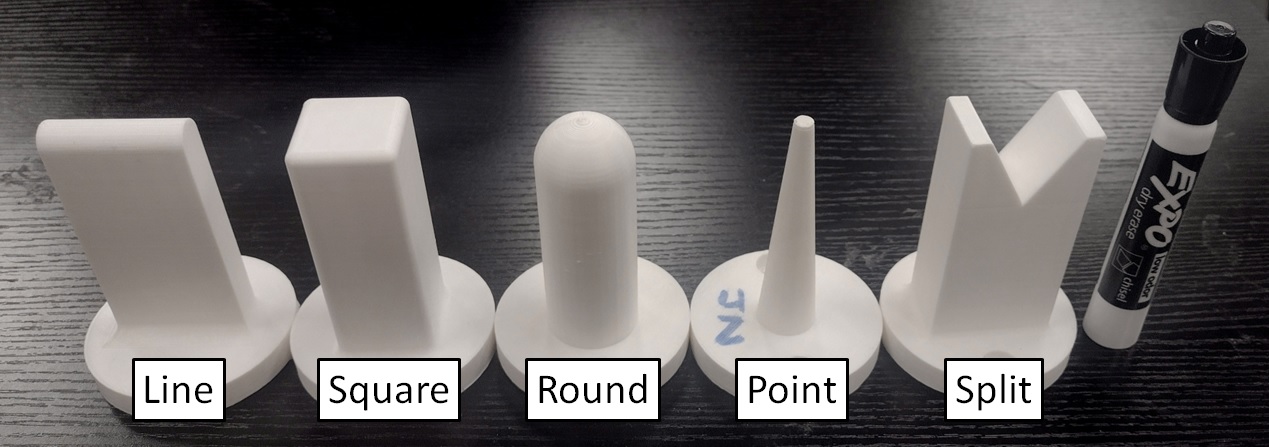}
    \caption{Photo of the experiment setup.
    A Punyo sensor is mounted to a UR5 robot, with a built-in force-torque sensor on its end effector.
    Indenters are mounted to a calibrated, fixed location.
    The 5 indenters are shown with an EXPO marker for scale.
    The model was calibrated on 5 trajectories against the Round indenter and one trajectory against the Point indenter, and tested against all five indenters with varying trajectories.
    }
    \label{fig:experiment_procedure}
    \vspace{-0.2in}
\end{figure}

\begin{figure*}[htbp]
    \centering
    \begin{tabu}{X[l, 0.5] | X[c, 1] X[c, 1] X[c, 1] X[c, 1] X[c, 1.4]}
        \toprule
        {\begin{minipage}{\linewidth}
            {Indenter}
            
            \vspace{0.3in}
        \end{minipage}}
        & \includegraphics[width=0.3\linewidth]{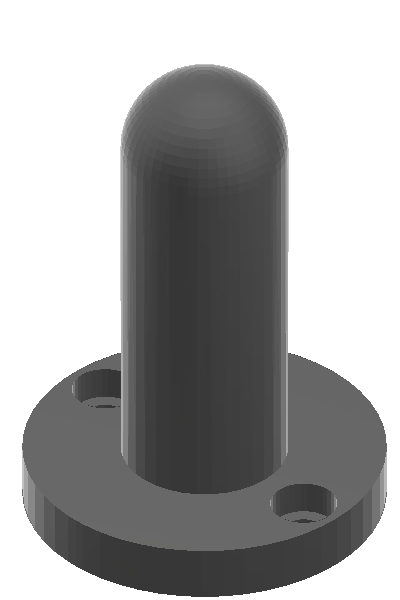}
        & \includegraphics[width=0.3\linewidth]{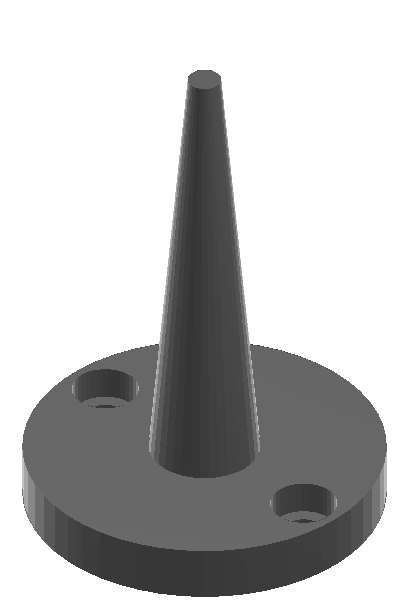}
        & \includegraphics[width=0.3\linewidth]{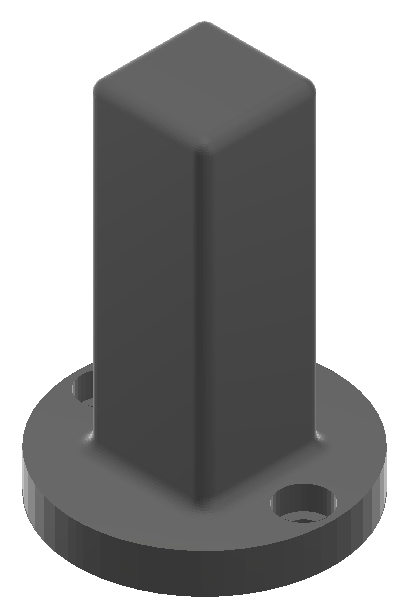}
        & \includegraphics[width=0.3\linewidth]{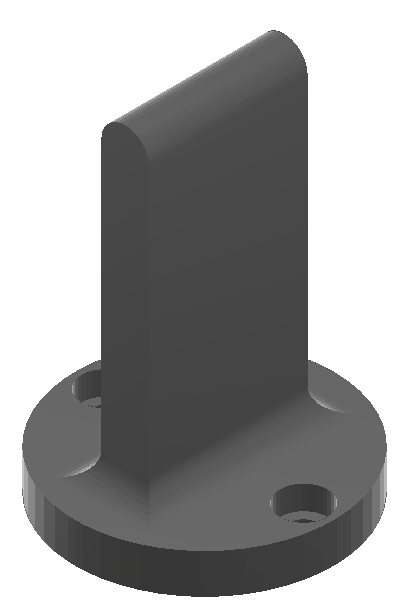}
        & {\begin{minipage}{\linewidth}
        \vspace{-0.4in}
        \hspace{0.3in}
        \includegraphics[width=0.23\linewidth]{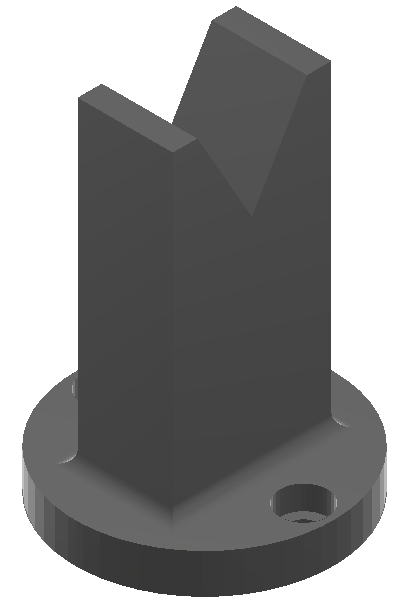}
        \end{minipage}}
        \\
        {\begin{minipage}{\linewidth}
            \vspace{0.3in}
        
            {Contact\\Estimate}
            
            \vspace{0.4in}
        \end{minipage}}
        &
        {\begin{minipage}{\linewidth}
        \vspace{-0.1in}
        \includegraphics[width=\linewidth]{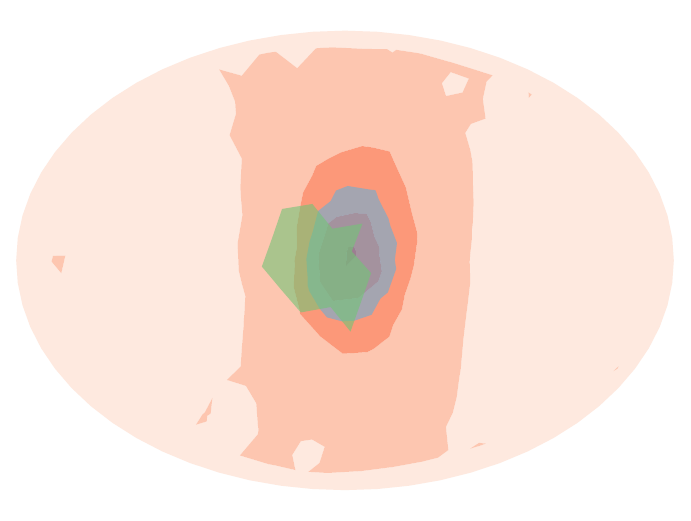}
        \end{minipage}}
        &
        {\begin{minipage}{\linewidth}
        \vspace{-0.1in}
        \includegraphics[width=\linewidth]{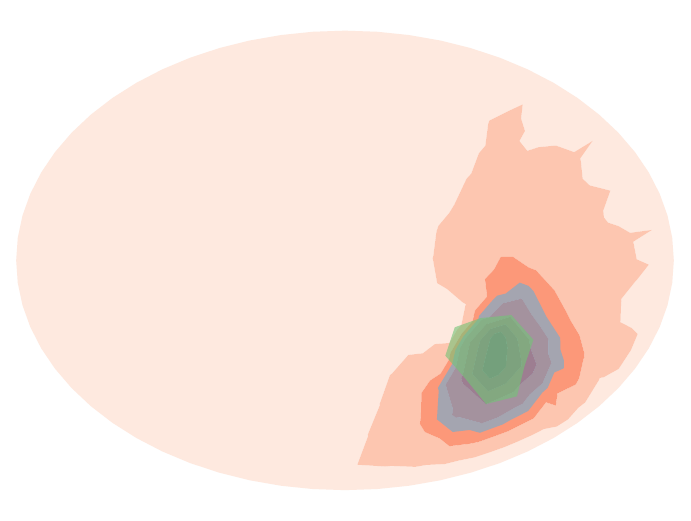}
        \end{minipage}}
        &
        {\begin{minipage}{\linewidth}
        \vspace{-0.1in}
        \includegraphics[width=\linewidth]{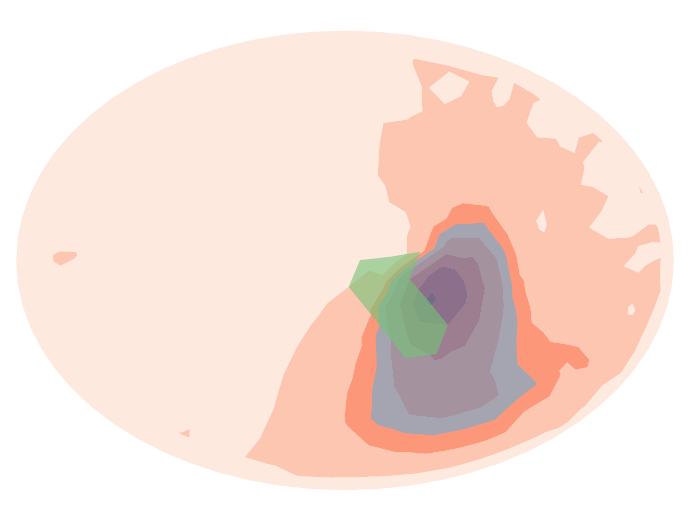}
        \end{minipage}}
        &
        {\begin{minipage}{\linewidth}
        \vspace{-0.1in}
        \includegraphics[width=\linewidth]{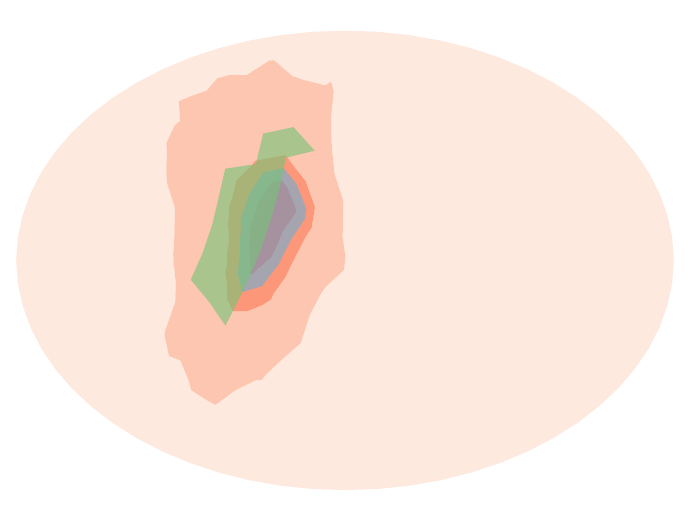}
        \end{minipage}}
        &
        {\begin{minipage}{\linewidth}
        \vspace{-0.1in}
        \includegraphics[width=\linewidth]{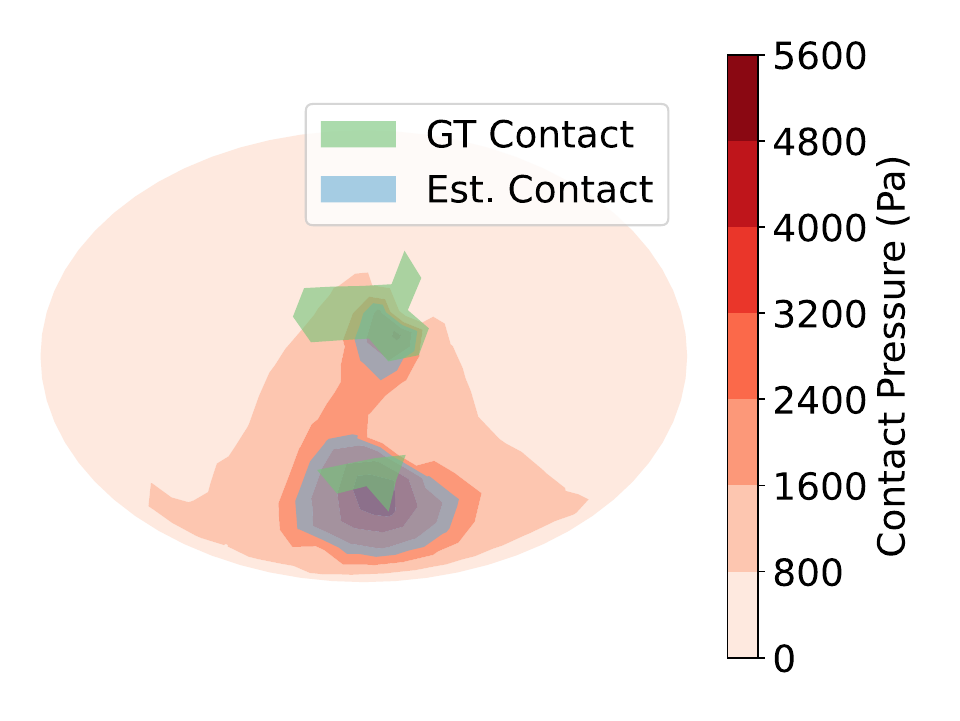}
        \end{minipage}}
        \\
        {\begin{minipage}{\linewidth}
            \vspace{-0.5in}
            
            {Force\\Field}
            
            \vspace{0.3in}
        \end{minipage}}
        & \includegraphics[width=\linewidth]{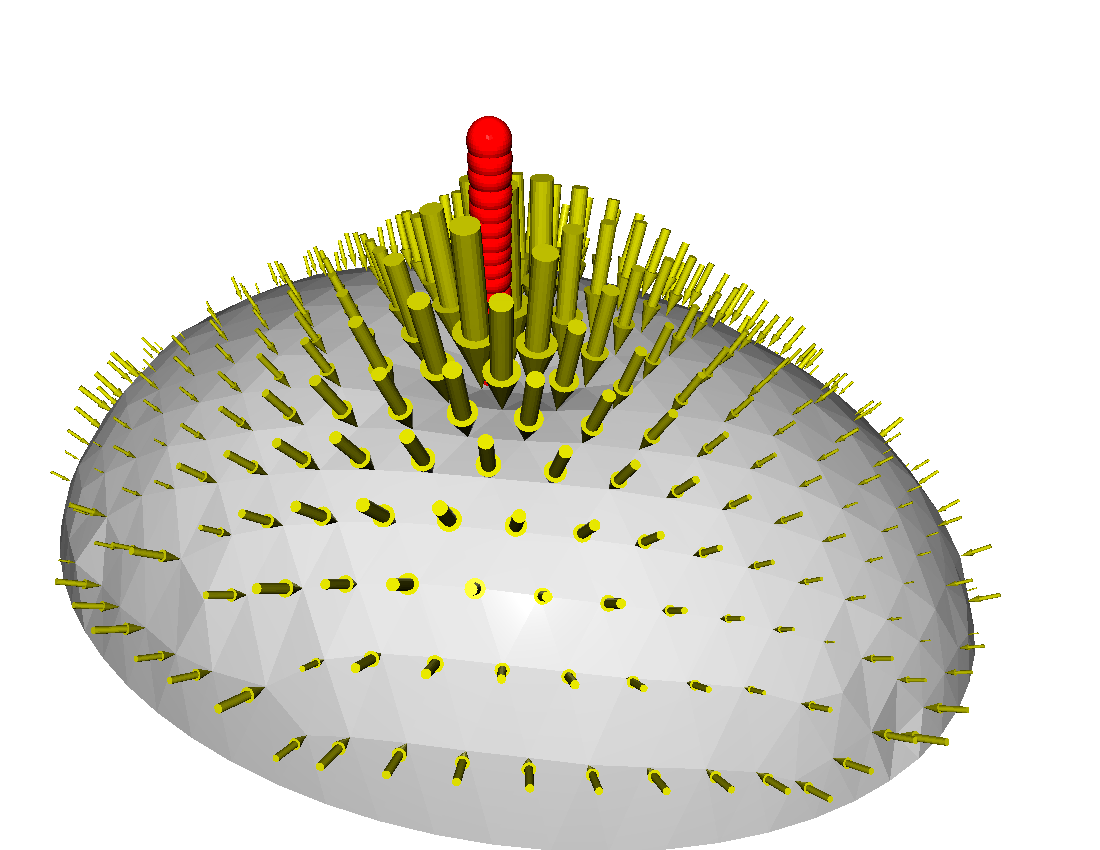}
        & \includegraphics[width=\linewidth]{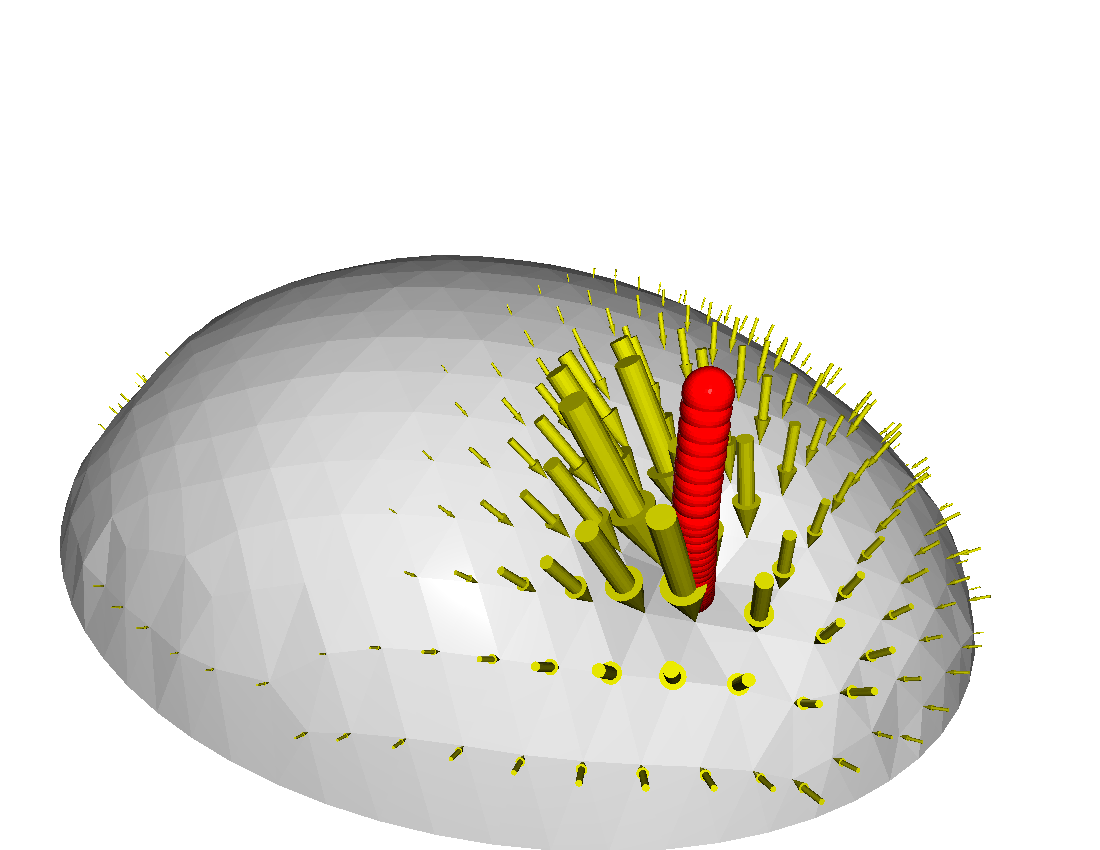}
        & \includegraphics[width=\linewidth]{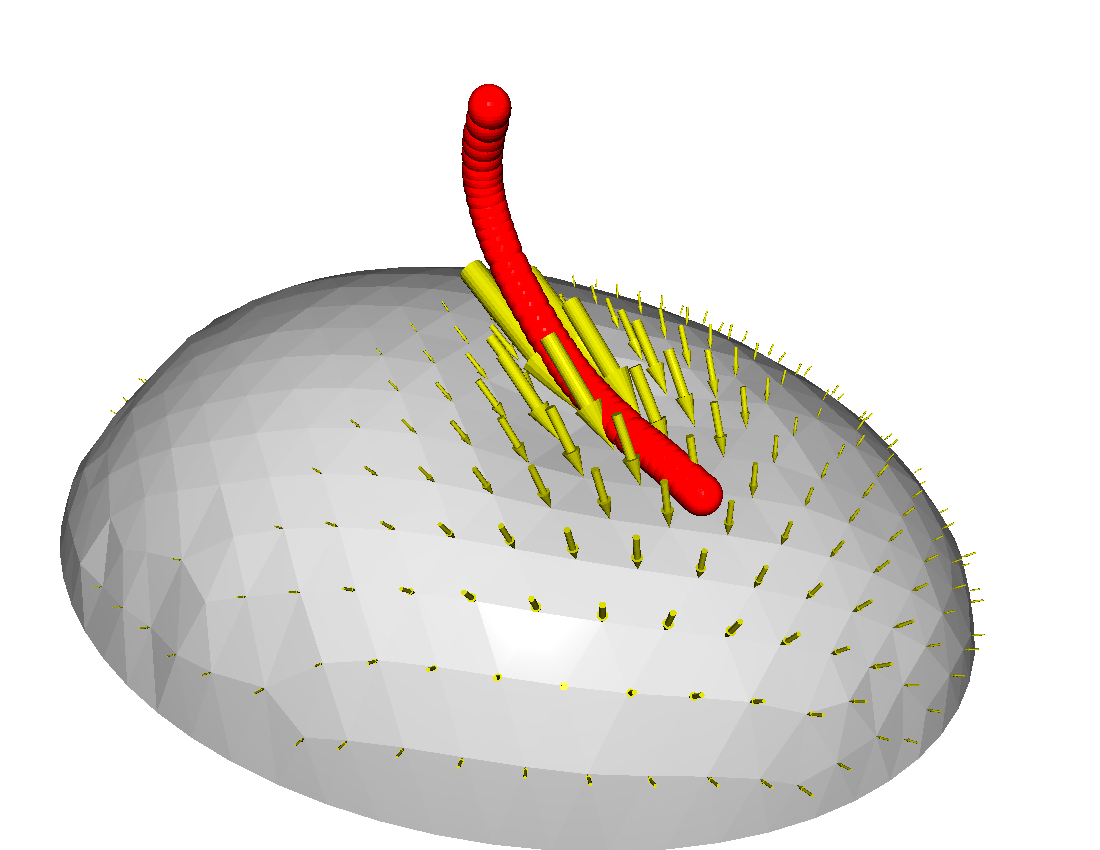}
        & \includegraphics[width=\linewidth]{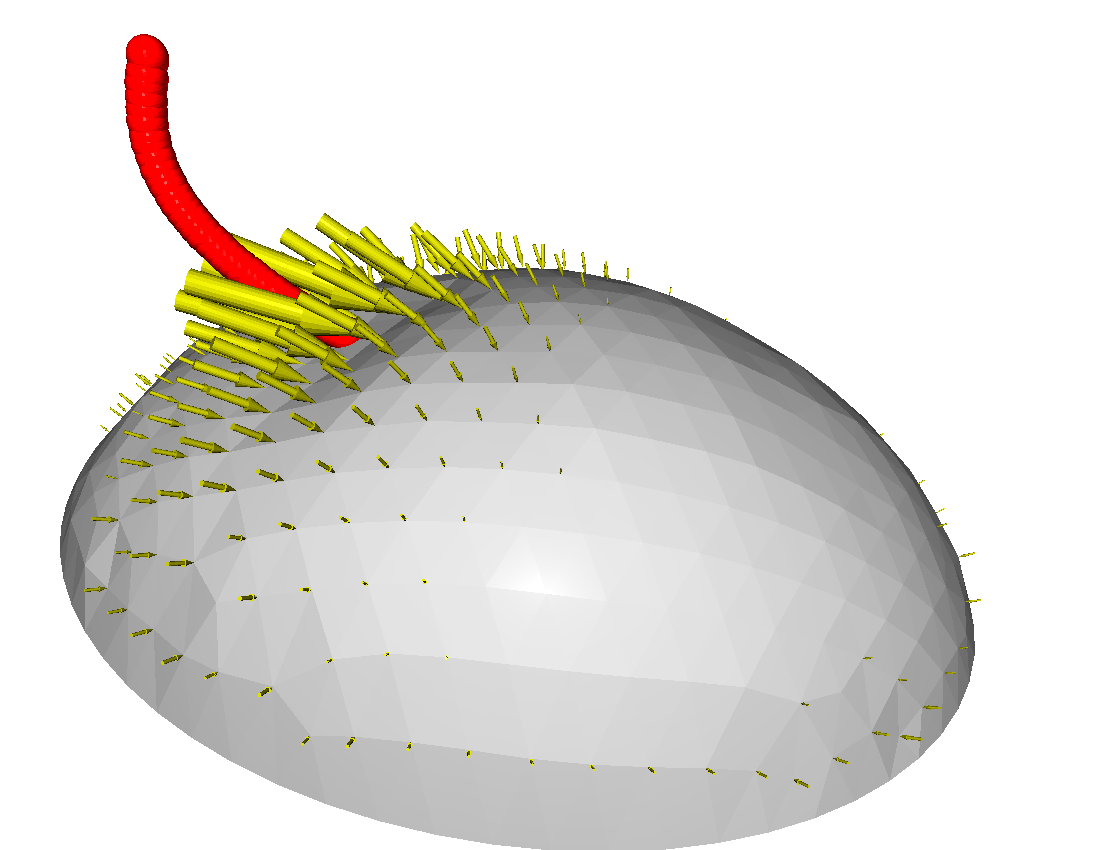}
        & {\begin{minipage}{\linewidth}
        \vspace{-0.8in}
        \includegraphics[width=0.77\linewidth]{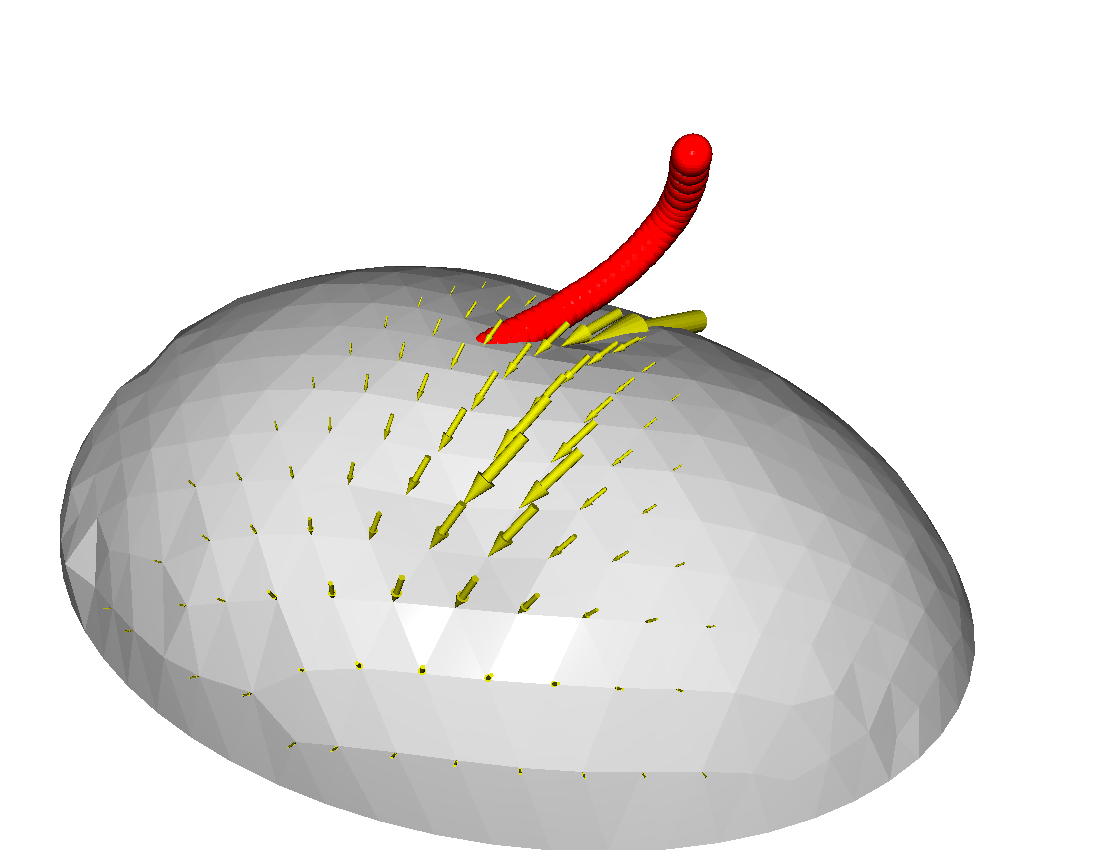}
        \end{minipage}}
        \\
        \bottomrule
    \end{tabu}
    \caption{Qualitative results of model predictions across different indenters and movement trajectories.
    In contact patch plots, the green region represents the ground truth contact, while the blue region represents the estimated contact patch.
    The contact trajectory is shown in red. [Best viewed in color]}
    \label{fig:many_results}
    \vspace{-0.2in}
\end{figure*}

\section{Model Calibration}
\label{sec:model_fitting}

Calibration of our model involves calibrating the material parameters $E, \nu$ and the optimization weight matrices $W_f, W_u$.
To reduce model complexity we set $W_f$ and $W_u$ as diagonal matrices.
We specify $k_f$ and $k_x$ as the penalties on the force and displacement terms respectively, and define
\begin{equation}
    \begin{split}
        (W_f)_{ii} & = \begin{cases}
            k_f \frac{\abs{\partial M}}{|M|} & \text{ node } i \in \partial M \\
            k_f & \text{ otherwise},
        \end{cases}\\
    \end{split}
\end{equation}
\begin{equation}
    (W_u)_{ii} = \frac{k_x}{\abs{M}}.
\end{equation}
This parametrization ensures that the magnitude of the force and displacement penalty terms are invariant to mesh element size:
To account for the faster growth of surface mesh points compared to boundary points as the mesh becomes more refined, the boundary force penalty must decrease to ensure the total force penalty stays constant.

We also fix the Poisson ratio $\nu$ at 0.5, modeling the rubber as incompressible.
This leaves the model with three parameters to calibrate:
$k_f$, $k_u$, and the Young's modulus $E$.

To calibrate the model, we collected a set of six trajectories of the robot pushing the sensor against an indenter (See \autoref{fig:experiment_procedure}).
Five trajectories were collected with the \texttt{round} indenter, and one with the \texttt{point} indenter.
Each trajectory consists of 30-50 data points, for a total of 205 data points.
Each data point consists of an RGB image, point cloud, pressure reading, and ground truth contact information.
We use known indentor geometry and the robot's kinematics to detect group truth contact regions.  
A force/torque sensor mounted on the UR5's wrist measures the ground truth contact force.
Note that in real-world applications, a robot is equipped with just the Punyo sensor, without a separate force/torque sensor.
The model's hyperparameters were optimized by blackbox function optimization~\cite{2020SciPy-NMeth}.

The optimization loss function is a function of the observation $O = (\text{RGB image, point cloud, } \delta p)$, and the ground truth force and contact measurements $\vec{f}_{gt}$ and $\vec{c}_{gt}$, where $c_{i,gt} = 1$ if node $i$ is in contact:
\begin{equation}
    \begin{split}
        L(O, \vec{f}_{gt}, \vec{c}_{gt}) & = 
        \lambda_1 \norm[2]{\vec{f}_{net}\p{O}- \vec{f}_{gt}} \\
            & + \lambda_2 \frac{1}{\abs{M}}\sum_{i=1}^{|M|}\norm[1]{C_i(O)  - c_{i,gt}} \\
        & \hspace{-0.7in} C_i(O) = \text{sigmoid}\p{k_{opt}\frac{p_i(O) - \text{thresh}(\vec{p}(O))}{\text{thresh}(\vec{p}(O))}}
    \end{split}
\end{equation}
where $\text{sigmoid}(x) = 1/(1+e^{-x})$.
$\lambda_1, \lambda_2, $ and $k_{opt}$ control the optimizer's penalty on force predictions, contact predictions, and contact prediction aggressiveness respectively.
We picked $\lambda_1 = 1 \text{ N}^{-1}, \lambda_2 = 10, $ and $k_{opt} = 2$ without tuning and they happened to work well.
We optimize the sum of the loss across all observations simultaneously.
The resulting optimized values for our set of trajectories are: $k_f = 0.3322 \text{ N}^{-1}, k_u = 537592 \text{ m}^{-2}, E = 1.317 \text{ MPa}$.

\section{Experiments}

We conduct experiments to evaluate the force and contact estimation accuracy of our proposed method.
We use the set of 5 indentors in~\autoref{fig:experiment_procedure} to generate a poking dataset, collecting ground truth contact regions and force measurements following the method in \autoref{sec:model_fitting}.
We compare our method against Kuppuswamy et al.~\cite{punyo_force} and demonstrate more accurate force prediction with comparable contact patch estimation.

\subsection{Dataset Preparation}

We sample poking trajectories with diverse locations and contact directions using the following procedure:
\begin{enumerate}
    \item Waypoint sampling: We sequentially sample three points. One in free space, one near the indenter, and one causing contact between the indenter and the sensor.  
    \item Sensor direction sampling: We sample a 3D orientation of the sensor which keeps it roughly pointed towards the indentor.
    We use the same orientation for the entire trajectory.  
    \item Trajectory validation: Given sampled waypoints and the sensor's target orientation, we use inverse kinematics to check the feasibility of the trajectory.
\end{enumerate}
The center of inflated membrane is used as the reference point to move along each sampled trajectory.

\begin{figure}
    \centering
    \includegraphics[width=0.49\linewidth]{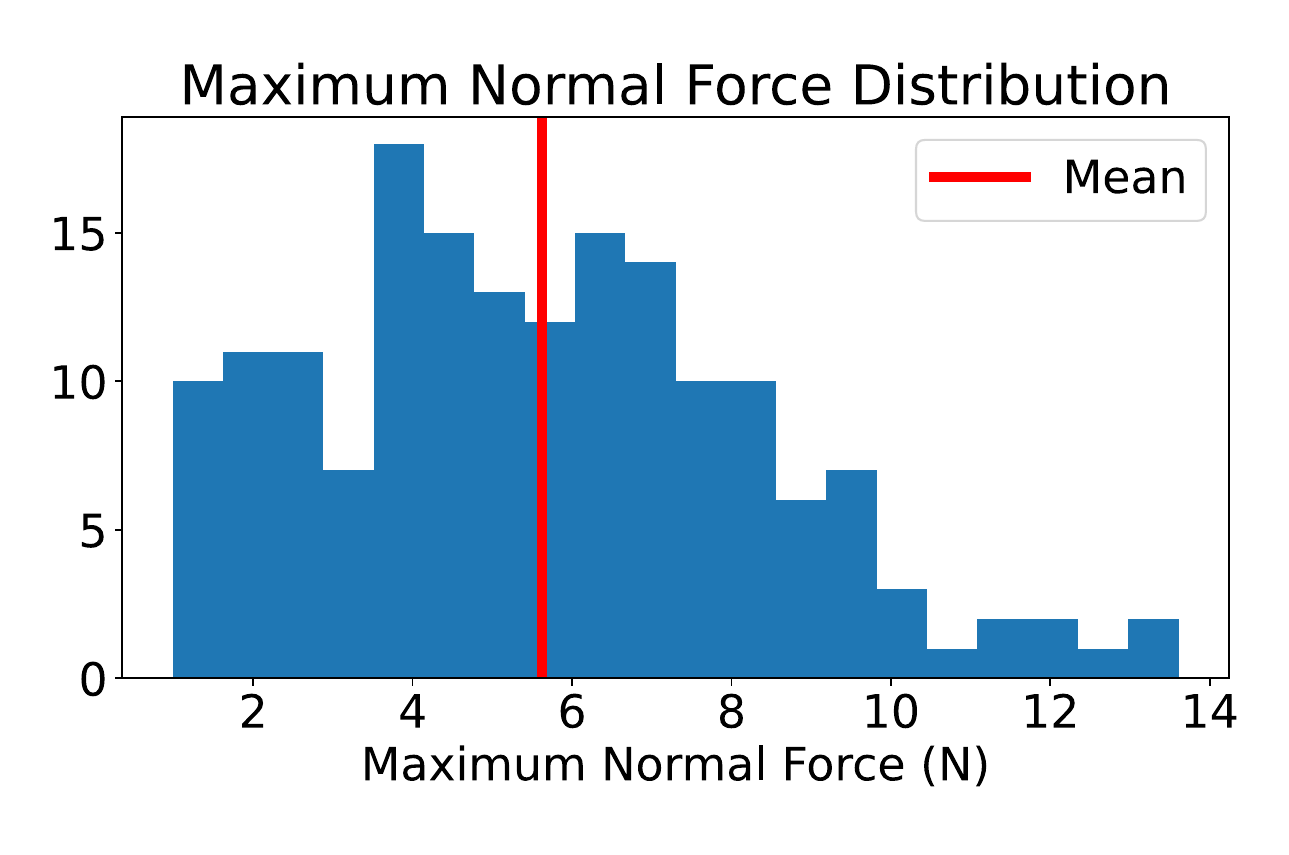}
    \includegraphics[width=0.49\linewidth]{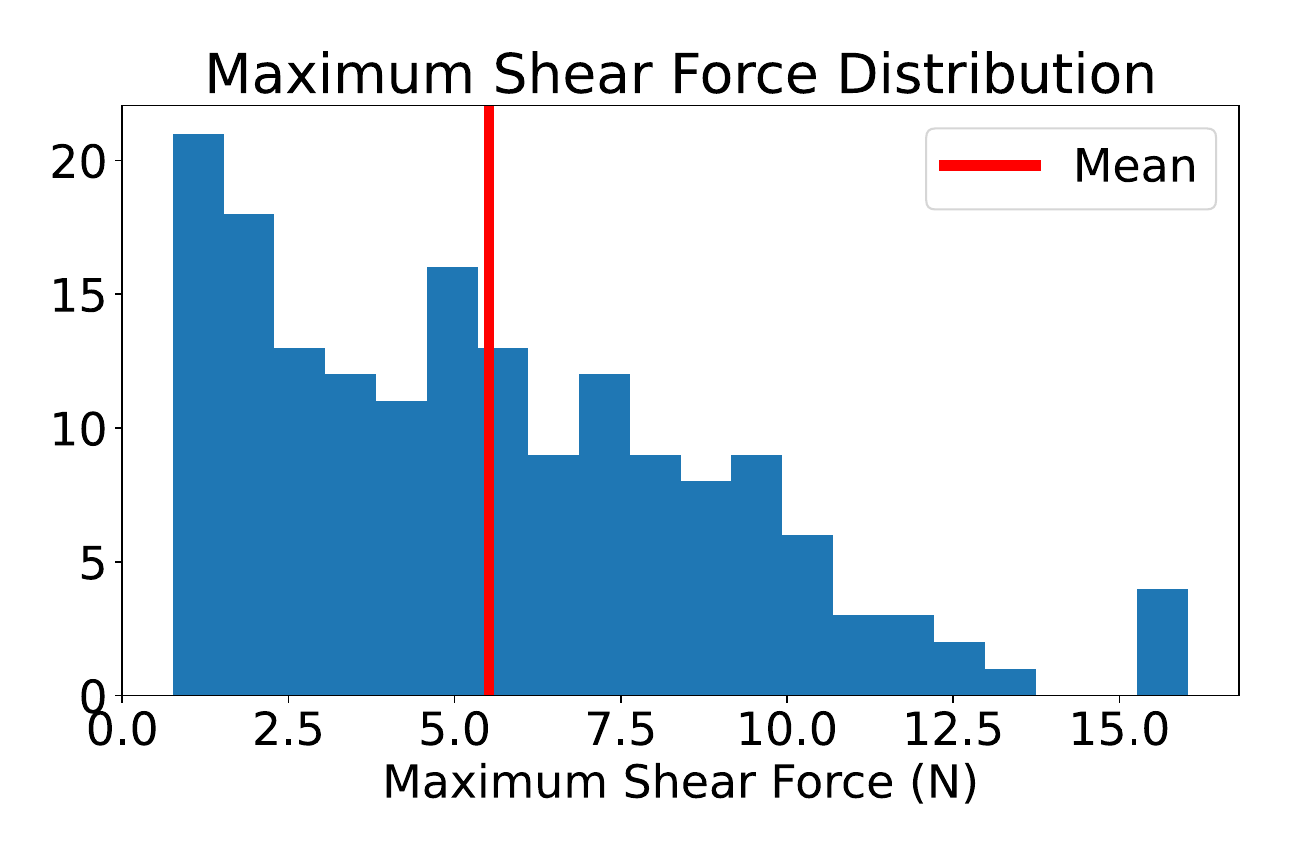}
    \includegraphics[width=0.49\linewidth]{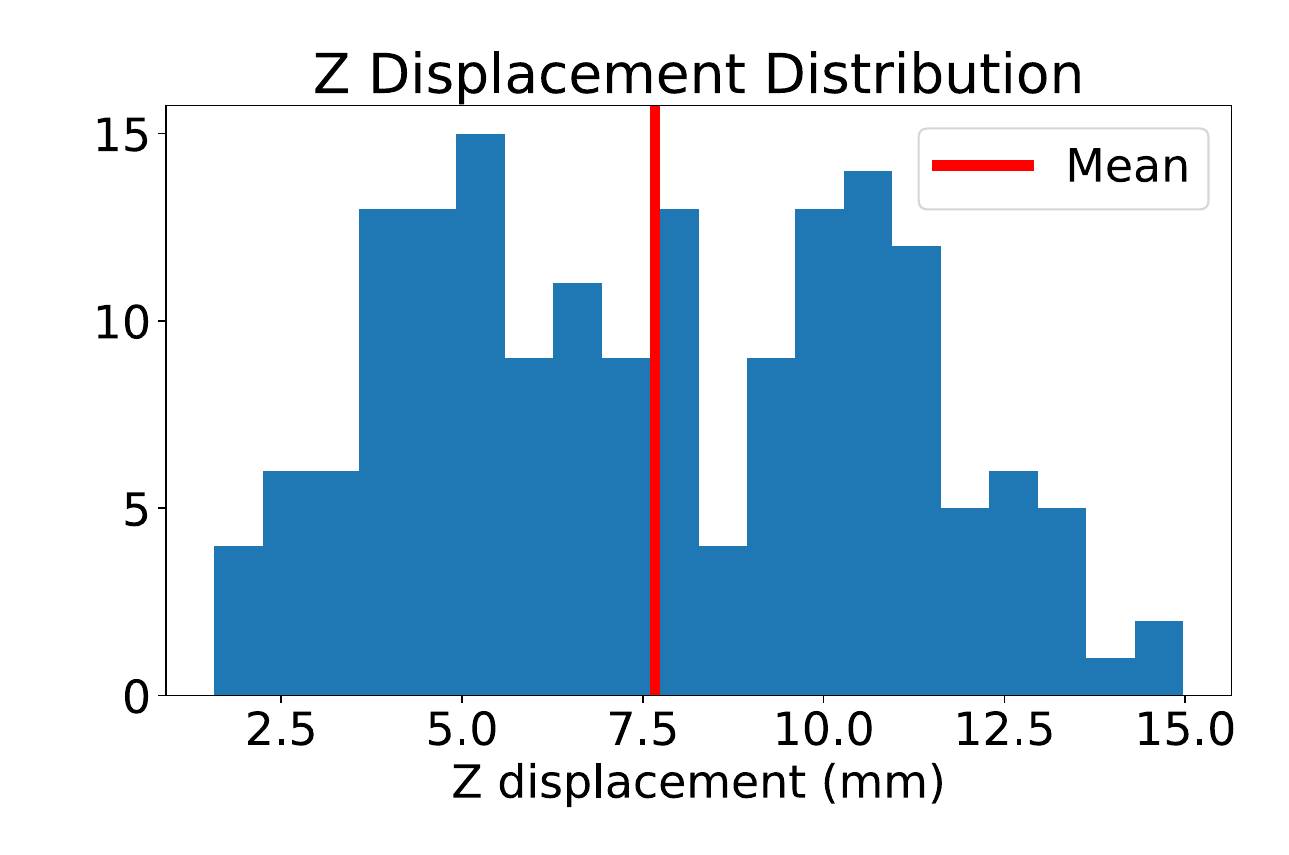}
    \includegraphics[width=0.49\linewidth]{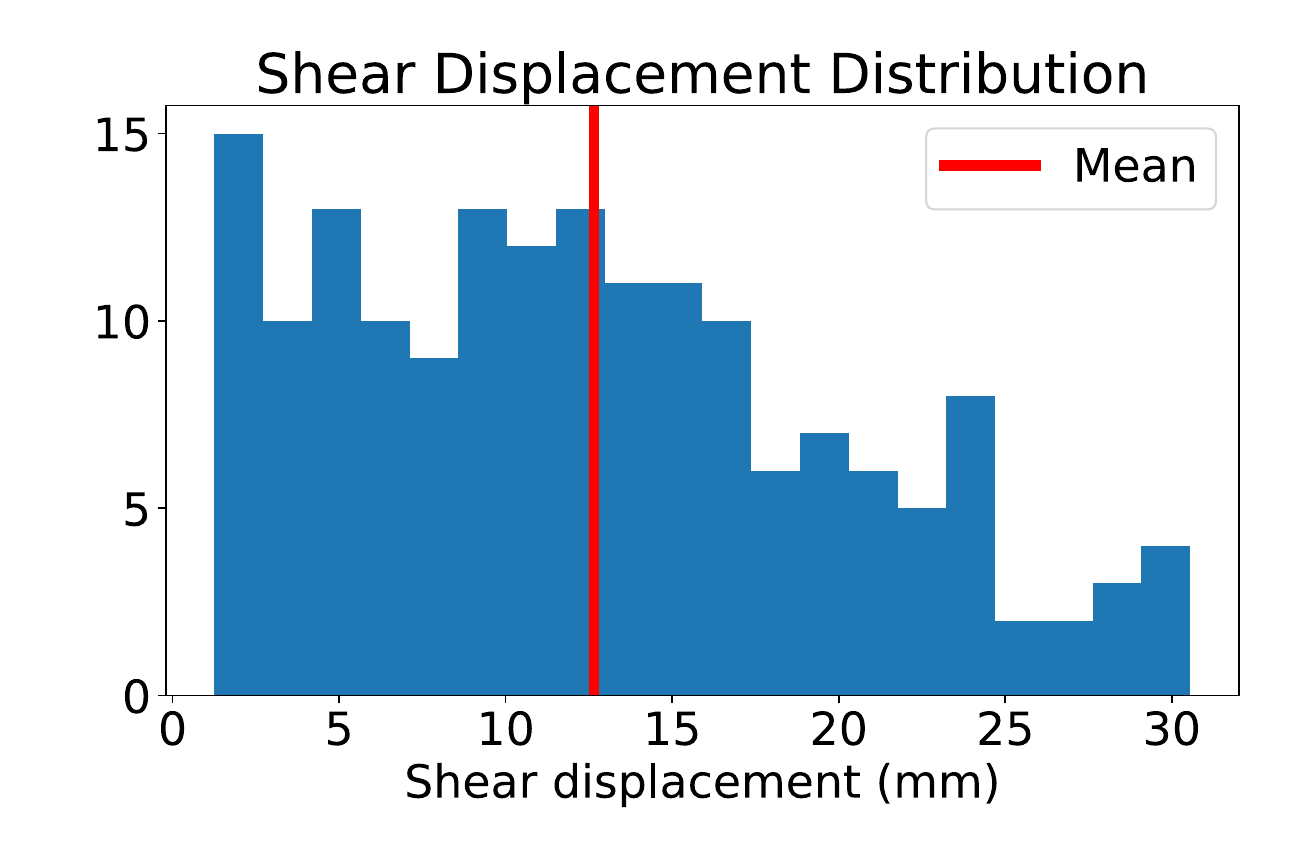}
    \caption{Summary of collected trajectory data.
    Collected trajectories cover light and heavy touches, with varying shear and normal components.}
    \label{fig:traj_summary}
\end{figure}

We executed 34 sampled trajectories against each of five different indenter shapes, for a total of 170 testing trajectories.
Each trajectory consists of 100 data points, as described in \autoref{sec:model_fitting}.
A summary of the collected trajectories is given in \autoref{fig:traj_summary}, showing that they are fairly diverse and cover a large range of displacements and contact forces in both normal and shear dimensions.  

\begin{figure}
    \includegraphics[width=\linewidth]{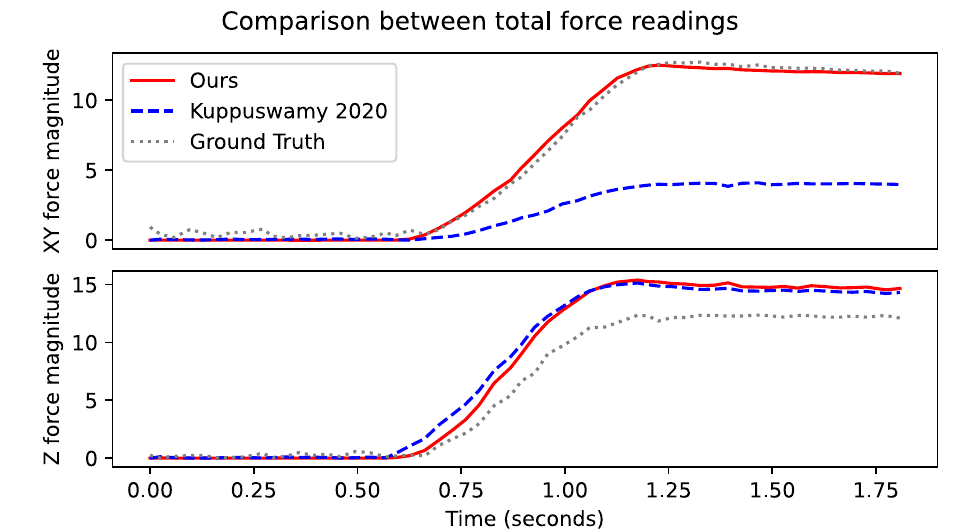}
    \includegraphics[width=\linewidth]{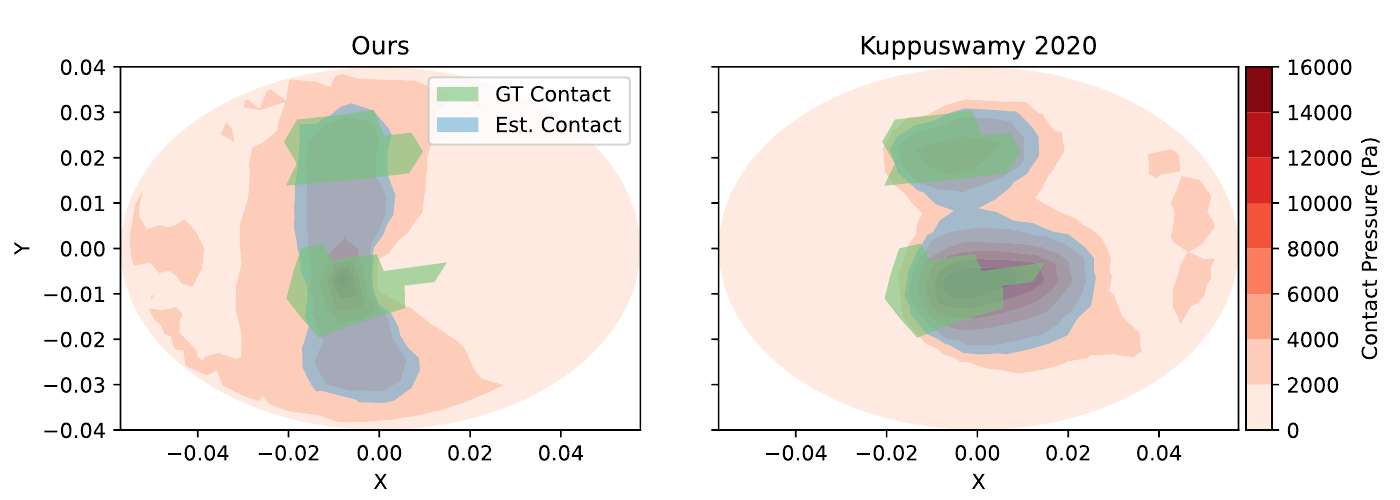}
    \caption{Qualitative comparison between our model and \cite{punyo_force}.
    Our model predicts more accurate XY forces due to accounting for shear effects.
    However, compared to using curvature information, our model is worse at distinguishing separate contact regions. [Best viewed in color]
    }
    \label{fig:force_qualitative}
    \vspace{-0.2in}
\end{figure}

\subsection{Accuracy Evaluation}

Results are shown in \autoref{tab:quantatative results} and \autoref{fig:many_results}, comparing our model to the baseline frictionless FE model of Kuppuswamy et al.~\cite{punyo_force}.  Our model is able to produce accurate total contact force predictions both qualitatively and quantitatively.
\autoref{fig:force_qualitative} show contact force prediction results aligned with the ground truth contact forces.
We significantly reduce the force prediction error by 36\% on average because our model can capture shear forces well.
Our method is able to predict force distributions that reflect the general nature of the contact, as seen in \autoref{fig:many_results}.
The choice of $L_1$ regularization lets the optimization more easily reject noise introduced by the sensor, but also tends to reject weaker touches: Our model tends to underestimate or fail to detect lighter contacts (net contact force $\sim 1$ N).
In addition, the linear plane stress approximation behaves very poorly in the presence of large curvature changes.
These effects result in slightly worse contact patch prediction accuracy compared to the baseline, resulting in a 10\% reduction in mIOU (visible in \autoref{fig:force_qualitative} and the \texttt{square} column of \autoref{fig:many_results}).


\subsection{Runtime and Scaling}

\begin{figure}
    \includegraphics[width=\linewidth]{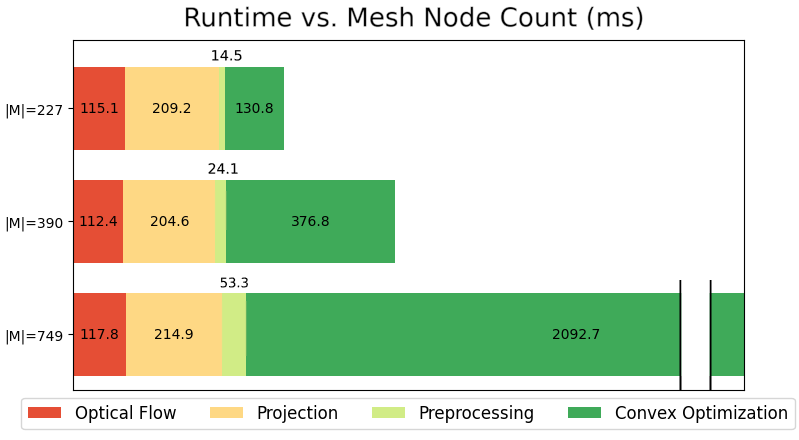}
    \caption{
    Average running time breakdown of our model at different mesh sizes.
    At coarser discretizations the model runs in near real time ($\sim2$ Hz).
    The time taken to solve the convex optimization problem scales quickly with the number of mesh nodes.
    [Best viewed in color]
    }
    \label{fig:runtime}
\end{figure}

Our definition of $k_f$ and $k_w$ allows the same set of tuned constants to be used with different mesh resolutions.
We tested three different mesh resolutions, coarse ($|M|=227$), medium ($|M|=390$), and fine ($|M|=749$), shown in \autoref{fig:runtime}.
Our pipeline runs in near-realtime with the coarse or medium mesh ($1-2$ Hz), while the fine mesh requires a $\sim{4}\times$ increase in computation time.

We observe the accuracy of our model does not change significantly as mesh resolution increases, as the total predicted force stays consistent.
The predicted contact area tends to behave similarly, with only small gains in resolution.
Hence, the data and figures in this paper reflect the outputs from the medium mesh ($\abs{M} = 390$).

\begin{table}[tbp]
    \centering
    \caption{Comparison of force and contact predictions. Our method performs worse than the baseline at contact detection but gives significantly more accurate force estimation.}
    \label{tab:quantatative results}
    \begin{tabu}{X[l] X[l] X[l] X[l] X[l]}
        \toprule
        & \multicolumn{2}{c}{Ours} &  \multicolumn{2}{c}{Kuppuswamy et al.~\cite{punyo_force}} \\
        Geometry & Contact mIOU $\uparrow$ & Avg. Force error (N) $\downarrow$ & Contact mIOU $\uparrow$ & Avg. Force error (N) $\downarrow$ \\
        \midrule
        Round & 0.528 & \textbf{1.04} & \textbf{0.597} & 1.34 \\
        Point & 0.530 & \textbf{1.16} & \textbf{0.555} & 1.77 \\
        Line & 0.542 & \textbf{1.18} & \textbf{0.574} & 1.81 \\
        Square & 0.485 & \textbf{1.33} & \textbf{0.540} & 2.25 \\
        Split & 0.478 & \textbf{1.52} & \textbf{0.537} & 2.56 \\
        \midrule
        Average & 0.513 & \textbf{1.24} & \textbf{0.560} & 1.95 \\
        \bottomrule
    \end{tabu}
    \vspace{-0.2in}
\end{table}


\section{Conclusion}

We presented a finite element force estimation method for soft-bubble grippers with only three parameters that can be calibrated with small amounts of data.
Our model can run in near real-time and produce force predictions with accuracy beyond the current state of the art, especially for shear forces.
In future work, we hope to develop a more accurate physical model for the bubble's deformation:
Our current bubble model uses a simplified model of membrane deformations, ignoring any effect from changes in the bubble's curvature or from large displacements changing the orientation of individual mesh elements.
Higher order elements including curvature effects should also improve accuracy.
We also hope to achieve speed improvements by implementation in a compiled language. 

\section{Acknowledgements}
We would like to thank Patrick Naughton, James Nam, Yifan Zhu, and Mengchao Zhang for help in proofreading and editing this paper.  We thank Toyota Research Institute for loaning us the Punyo soft-bubble sensor.

\printbibliography

@string{ICRA="IEEE Int. Conf. Robotics and Automation"}

@Article{sensors2014,
    AUTHOR = {Ito, Yuji and Kim, Youngwoo and Obinata, Goro},
    TITLE = {Contact Region Estimation Based on a Vision-Based Tactile Sensor Using a Deformable Touchpad},
    JOURNAL = {Sensors},
    VOLUME = {14},
    YEAR = {2014},
    NUMBER = {4},
    PAGES = {5805--5822},
    URL = {https://www.mdpi.com/1424-8220/14/4/5805},
    PubMedID = {24670719},
    ISSN = {1424-8220},
    DOI = {10.3390/s140405805}
}

@Article{sensors2018,
    AUTHOR = {Chi, Cheng and Sun, Xuguang and Xue, Ning and Li, Tong and Liu, Chang},
    TITLE = {Recent Progress in Technologies for Tactile Sensors},
    JOURNAL = {Sensors},
    VOLUME = {18},
    YEAR = {2018},
    NUMBER = {4},
    ARTICLE-NUMBER = {948},
    URL = {https://www.mdpi.com/1424-8220/18/4/948},
    PubMedID = {29565835},
    ISSN = {1424-8220},
    DOI = {10.3390/s18040948}
}

@Article{Shah2021,
    AUTHOR = {Shah, Umer Hameed and Muthusamy, Rajkumar and Gan, Dongming and Zweiri, Yahya and Seneviratne, Lakmal},
    TITLE = {On the Design and Development of Vision-based Tactile Sensors},
    JOURNAL = {Journal of Intelligent \& Robotic Systems},
    VOLUME = {102},
    YEAR = {2021},
    NUMBER = {82},
    URL = {https://doi.org/10.1007/s10846-021-01431-0},
    ISSN = {1573-0409},
    DOI = {10.1007/s10846-021-01431-0}
}

@Article{gelsight,
    AUTHOR = {Yuan, Wenzhen and Dong, Siyuan and Adelson, Edward H.},
    TITLE = {GelSight: High-Resolution Robot Tactile Sensors for Estimating Geometry and Force},
    JOURNAL = {Sensors},
    VOLUME = {17},
    YEAR = {2017},
    NUMBER = {12},
    ARTICLE-NUMBER = {2762},
    URL = {https://www.mdpi.com/1424-8220/17/12/2762},
    PubMedID = {29186053},
    ISSN = {1424-8220},
    DOI = {10.3390/s17122762}
}

@InProceedings{oller2023corl,
      title = 	 {Manipulation via Membranes: High-Resolution and Highly Deformable Tactile Sensing and Control},
      author =       {Oller, Miquel and Lisbona, Mireia Planas i and Berenson, Dmitry and Fazeli, Nima},
      booktitle = 	 {Proceedings of The 6th Conference on Robot Learning},
      pages = 	 {1850--1859},
      year = 	 {2023},
      editor = 	 {Liu, Karen and Kulic, Dana and Ichnowski, Jeff},
      volume = 	 {205},
      series = 	 {Proceedings of Machine Learning Research},
      month = 	 {14--18 Dec},
      publisher =    {PMLR},
      url = 	 {https://proceedings.mlr.press/v205/oller23a.html}
}

@article{lambeta2020digit,
  title={Digit: A novel design for a low-cost compact high-resolution tactile sensor with application to in-hand manipulation},
  author={Lambeta, Mike and Chou, Po-Wei and Tian, Stephen and Yang, Brian and Maloon, Benjamin and Most, Victoria Rose and Stroud, Dave and Santos, Raymond and Byagowi, Ahmad and Kammerer, Gregg and others},
  journal={IEEE Robotics and Automation Letters},
  volume={5},
  number={3},
  pages={3838--3845},
  year={2020},
  publisher={IEEE}
}

@article{tactip,
    author = {Ward-Cherrier, Benjamin and Pestell, Nicholas and Cramphorn, Luke and Winstone, Benjamin and Giannaccini, Maria and Rossiter, Jonathan and Lepora, Nathan},
    year = {2018},
    month = {01},
    pages = {},
    title = {The TacTip Family: Soft Optical Tactile Sensors with 3D-Printed Biomimetic Morphologies},
    volume = {5},
    journal = {Soft Robotics},
    doi = {10.1089/soro.2017.0052}
}

@article{alspach2019softbubble,
  title={Soft-bubble: A highly compliant dense geometry tactile sensor for robot manipulation},
  author={Alexander Alspach and Kunimatsu Hashimoto and Naveen Suresh Kuppuswamy and Russ Tedrake},
  journal={2019 2nd IEEE International Conference on Soft Robotics (RoboSoft)},
  year={2019},
  pages={597-604},
  url={https://api.semanticscholar.org/CorpusID:102336499}
}

@article{Wang2021GelSightWM,
  title={GelSight Wedge: Measuring High-Resolution 3D Contact Geometry with a Compact Robot Finger},
  author={Shaoxiong Wang and Yu She and Branden Romero and Edward H. Adelson},
  journal={2021 IEEE International Conference on Robotics and Automation (ICRA)},
  year={2021},
  pages={6468-6475},
  url={https://api.semanticscholar.org/CorpusID:235446468}
}

@inproceedings{lambeta2021pytouch,
  title={PyTouch: A machine learning library for touch processing},
  author={Lambeta, Mike and Xu, Huazhe and Xu, Jingwei and Chou, Po-Wei and Wang, Shaoxiong and Darrell, Trevor and Calandra, Roberto},
  booktitle={2021 IEEE International Conference on Robotics and Automation (ICRA)},
  pages={13208--13214},
  year={2021},
  organization={IEEE}
}

@inproceedings{funk2023highresolution,
    title={High-Resolution Pixelwise Contact Area and Normal Force Estimation for the GelSight Mini Visuotactile Sensor Using Neural Networks},
    author={Niklas Funk and Paul Otto M{\"u}ller and Boris Belousov and Anton Savchenko and Rolf Findeisen and Jan Peters},
    booktitle={Embracing Contacts - Workshop at ICRA 2023},
    year={2023},
    url={https://openreview.net/forum?id=dUO0QQw4FW}
}

@ARTICLE{punyo_force,
  author={Kuppuswamy, Naveen and Castro, Alejandro and Phillips-Grafflin, Calder and Alspach, Alex and Tedrake, Russ},
  journal={IEEE Robotics and Automation Letters}, 
  title={Fast Model-Based Contact Patch and Pose Estimation for Highly Deformable Dense-Geometry Tactile Sensors}, 
  year={2020},
  volume={5},
  number={2},
  pages={1811-1818},
  doi={10.1109/LRA.2019.2961050}
}

@INPROCEEDINGS{gelsight_FEM,
  author={Ma, Daolin and Donlon, Elliott and Dong, Siyuan and Rodriguez, Alberto},
  booktitle={2019 International Conference on Robotics and Automation (ICRA)}, 
  title={Dense Tactile Force Estimation using GelSlim and inverse FEM}, 
  year={2019},
  volume={},
  number={},
  pages={5418-5424},
  doi={10.1109/ICRA.2019.8794113}
}

@InProceedings{farneback_flow,
    author="Farneb{\"a}ck, Gunnar",
    editor="Bigun, Josef
    and Gustavsson, Tomas",
    title="Two-Frame Motion Estimation Based on Polynomial Expansion",
    booktitle="Image Analysis",
    year="2003",
    publisher="Springer Berlin Heidelberg",
    address="Berlin, Heidelberg",
    pages="363--370",
    isbn="978-3-540-45103-7"
}

@article{opencv_library,
    author = {Bradski, G.},
    journal = {Dr. Dobb's Journal of Software Tools},
    title = {{The OpenCV Library}},
    year = {2000}
}

@ARTICLE{2020SciPy-NMeth,
  author  = {Virtanen, Pauli and Gommers, Ralf and Oliphant, Travis E. and
            Haberland, Matt and Reddy, Tyler and Cournapeau, David and
            Burovski, Evgeni and Peterson, Pearu and Weckesser, Warren and
            Bright, Jonathan and {van der Walt}, St{\'e}fan J. and
            Brett, Matthew and Wilson, Joshua and Millman, K. Jarrod and
            Mayorov, Nikolay and Nelson, Andrew R. J. and Jones, Eric and
            Kern, Robert and Larson, Eric and Carey, C J and
            Polat, {\.I}lhan and Feng, Yu and Moore, Eric W. and
            {VanderPlas}, Jake and Laxalde, Denis and Perktold, Josef and
            Cimrman, Robert and Henriksen, Ian and Quintero, E. A. and
            Harris, Charles R. and Archibald, Anne M. and
            Ribeiro, Ant{\^o}nio H. and Pedregosa, Fabian and
            {van Mulbregt}, Paul and {SciPy 1.0 Contributors}},
  title   = {{{SciPy} 1.0: Fundamental Algorithms for Scientific
            Computing in Python}},
  journal = {Nature Methods},
  year    = {2020},
  volume  = {17},
  pages   = {261--272},
  adsurl  = {https://rdcu.be/b08Wh},
  doi     = {10.1038/s41592-019-0686-2},
}

@book{fem_book,
author = {Hughes, Thomas J. R.},
address = {Mineola, NY},
booktitle = {The finite element method : linear static and dynamic finite element analysis},
isbn = {0486411818},
language = {eng},
lccn = {00038414},
publisher = {Dover Publications},
title = {The finite element method : linear static and dynamic finite element analysis },
year = {2000 - 1987},
}

@article{she2021ijrr,
    author = {Yu She and Shaoxiong Wang and Siyuan Dong and Neha Sunil and Alberto Rodriguez and Edward Adelson},
    title ={Cable manipulation with a tactile-reactive gripper},
    journal = {The International Journal of Robotics Research},
    volume = {40},
    number = {12-14},
    pages = {1385-1401},
    year = {2021},
    doi = {10.1177/02783649211027233},
    URL = {https://doi.org/10.1177/02783649211027233},
    eprint = {https://doi.org/10.1177/02783649211027233}
}

@INPROCEEDINGS{tang2023rss, 
    AUTHOR    = {Bingjie Tang AND Michael A Lin AND Iretiayo A Akinola AND Ankur Handa AND Gaurav S Sukhatme AND Fabio Ramos AND Dieter Fox AND Yashraj S Narang}, 
    TITLE     = {{IndustReal: Transferring Contact-Rich Assembly Tasks from Simulation to Reality}}, 
    BOOKTITLE = {Proceedings of Robotics: Science and Systems}, 
    YEAR      = {2023}, 
    ADDRESS   = {Daegu, Republic of Korea}, 
    MONTH     = {July}, 
    DOI       = {10.15607/RSS.2023.XIX.039} 
}

@INPROCEEDINGS{erichson2018icra,
  author={Erickson, Zackory and Clever, Henry M. and Turk, Greg and Liu, C. Karen and Kemp, Charles C.},
  booktitle={2018 IEEE International Conference on Robotics and Automation (ICRA)}, 
  title={Deep Haptic Model Predictive Control for Robot-Assisted Dressing}, 
  year={2018},
  volume={},
  number={},
  pages={4437-4444},
  doi={10.1109/ICRA.2018.8460656}
}

@INPROCEEDINGS{yuan2017icra,
  author={Yuan, Wenzhen and Zhu, Chenzhuo and Owens, Andrew and Srinivasan, Mandayam A. and Adelson, Edward H.},
  booktitle={2017 IEEE International Conference on Robotics and Automation (ICRA)}, 
  title={Shape-independent hardness estimation using deep learning and a GelSight tactile sensor}, 
  year={2017},
  volume={},
  number={},
  pages={951-958},
  doi={10.1109/ICRA.2017.7989116}
}

@article{diamond2016cvxpy,
  author  = {Steven Diamond and Stephen Boyd},
  title   = {{CVXPY}: {A} {P}ython-embedded modeling language for convex optimization},
  journal = {Journal of Machine Learning Research},
  year    = {2016},
  volume  = {17},
  number  = {83},
  pages   = {1--5},
}

@INPROCEEDINGS{do2023densetact,
  author={Do, Won Kyung and Jurewicz, Bianca and Kennedy, Monroe},
  booktitle={2023 IEEE International Conference on Robotics and Automation (ICRA)}, 
  title={DenseTact 2.0: Optical Tactile Sensor for Shape and Force Reconstruction}, 
  year={2023},
  volume={},
  number={},
  pages={12549-12555},
  doi={10.1109/ICRA48891.2023.10161150}
}

\end{document}